\documentclass[lettersize,journal]{IEEEtran}
\usepackage{amsmath,amsfonts}
\usepackage{algorithmic}
\usepackage{algorithm}
\usepackage{array}
\usepackage[caption=false,font=normalsize,labelfont=sf,textfont=sf]{subfig}
\usepackage{textcomp}
\usepackage{stfloats}
\usepackage{url}
\usepackage{verbatim}
\usepackage{graphicx}
\usepackage{cite}
\usepackage {subcaption}
\newcommand{\bfsection}[1]{\noindent\textbf{#1.}}
\newcommand{\model}{STGFormer}
\newcommand{\modelo}{STGFormer~}
\def\HS{\hspace{\fontdimen2\font}}
\newcommand{\red}[1]{\textcolor{red}{#1}}

\newcommand{\itp}{\textit{p}}
\newcommand{\itq}{\textit{q}}
\newcommand{\itx}{\textit{x}}
\newcommand{\bfx}{\textbf{x}}

\def\eg{\emph{e.g.}}

\usepackage[pagebackref=true,breaklinks=true,letterpaper=true,colorlinks,bookmarks=false]{hyperref}

\usepackage[capitalize]{cleveref}
\crefname{section}{Sec.}{Secs.}
\Crefname{section}{Section}{Sections}
\Crefname{table}{Table}{Tables}
\crefname{table}{Tab.}{Tabs.}

\begin{document}

\title{Learning Socio-Temporal Graphs for Multi-Agent Trajectory Prediction}

\author{Yuke Li, Lixiong Chen, Guangyi Chen, Ching-Yao Chan, Kun Zhang, Stefano Anzellotti, Donglai Wei
\thanks{Yuke Li and Donglai Wei are with the Computer Science Department, Boston College, 245 Beacon Street, Chestnut Hill, MA, 02135, USA (email: leesunfreshing@gmail.com). 

Lixiong Chen is with Department of Engineering Science, University of Oxford, Parks Road, Oxford, OX1 3PJ, UK.

Guangyi Chen and Kun Zhang are with Department of Philosophy, Carnegie Mellon University, 5000 Forbes Avenue Pittsburgh, PA 15213, USA.

Stefano Anzellotti is with Department of Psychology and NeuroScience, Boston College, 140 Commonwealth Avenue Chestnut Hill, MA 02467, USA.

Ching-Yao Chan is with California PATH, UC Berekeley, 1357 South 46th Street, Richmond, CA 94804, USA
}
\thanks{}}

\markboth{IEEE Transactions on Image Processing}%
{Shell \MakeLowercase{\textit{et al.}}: A Sample Article Using IEEEtran.cls for IEEE Journals}


\maketitle

\begin{abstract}
In order to predict a pedestrian's trajectory in a crowd accurately, one has to take into account her/his underlying socio-temporal interactions with other pedestrians consistently. Unlike existing work that represents the relevant information separately,  partially, or implicitly, we propose a complete representation for it to be fully and explicitly captured and analyzed. In particular, we introduce a Directed Acyclic Graph-based structure, which we term  \textbf{Socio-Temporal Graph} (STG), to explicitly capture pair-wise socio-temporal interactions among a group of people across both space and time. Our model is built on a time-varying generative process, whose latent variables determine the structure of the STGs. We design an attention-based model named \textbf{STGformer} that affords an end-to-end pipeline to learn the structure of the STGs for trajectory prediction. Our solution achieves overall state-of-the-art prediction accuracy in two large-scale benchmark datasets. Our analysis shows that a person's past trajectory is critical for predicting another person's future path. Our model learns this relationship with a strong notion of socio-temporal localities. Statistics show that utilizing this information explicitly for prediction yields a noticeable performance gain with respect to the trajectory-only approaches. 
\end{abstract}


\section{Introduction}
\label{sec:intro}


When a person walks in a crowd, in addition to the location of their destination, this person's motion path is also affected by other agents' movements, \eg following the lead, taking a detour to avoid collisions, or making a stop at a rendezvous. Accurately forecasting an agent's motion trajectories depends on how accurately their surrounding agents' activities are captured and modeled. In vision-based applications, including autonomous driving \cite{trajecpred_nips19,pred_traj_nips20,pred_cvpr21_1}, long-term object tracking \cite{trackpred_cvpr20}, and robotic planning \cite{pred_cvpr20,actpred_cvpr20}, the relevant information is often embedded in the relative motion between two agents, which is also echoed by findings in cognitive science \cite{jara2016naive,baker2017rational,kryven2021plans}.

\begin{figure}[t]
\centering

\includegraphics[width=\linewidth]{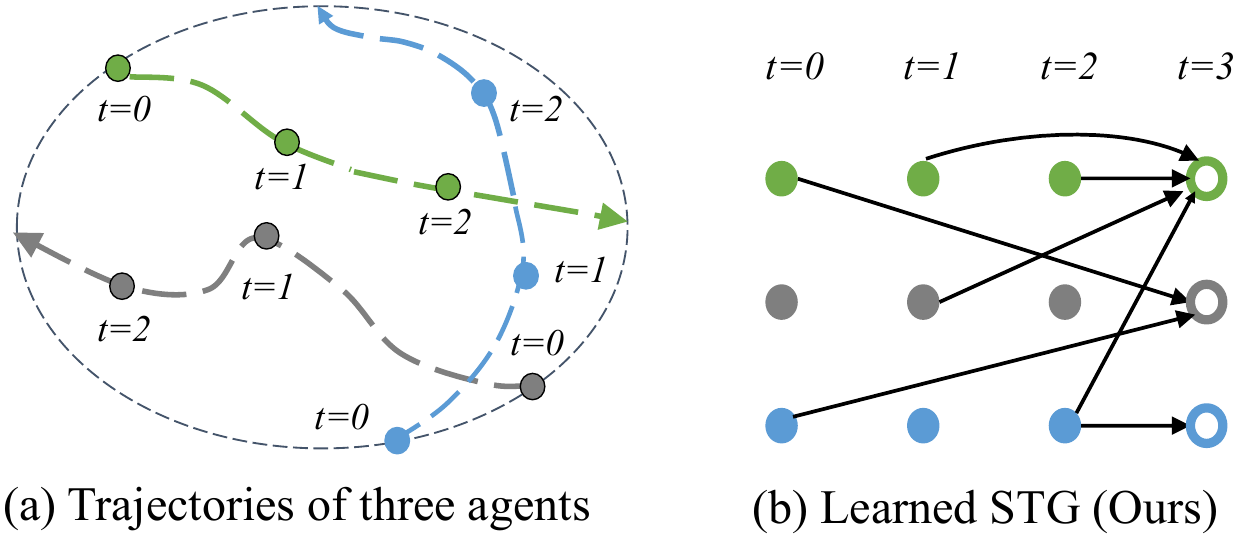}
\vspace{-0.6cm}
\caption{
(a) Given the observed trajectories of multiple agents at time $t=$0-2, (b) ations at $t-1$, the socio-temporal models for multi-agent trajectory forecastindg at $t$. (a) Most previous models assume a fixed structure, while (b) our model learns the socio-temporal graph (STG). }
\label{fig:teaser}
\vspace{-0.4cm}
\end{figure}


In a scene of only pedestrians, there exists a hidden graph that governs the socio-temporal interactions among the agents. Relative information is either discarded, partially represented, or implicitly encoded by various models proposed to date, such as pooling \cite{lstm_cvpr16,gan_cvpr18,trajecpred_4_eccv20}, social force \cite{pred_cvpr21_4,socialnce_cvpr21,pred_eccv22_2}, graph neural networks \cite{pred_traj_nips19,chen2021human}. Attention networks \cite{trajecpred_eccv20_1,agentformer_iccv21,pred_eccv22_1}. Even though without ground-truth labels, no restrictions are imposed upon predictive models in terms of what information should be utilized and how to represent it for prediction, for accurate and robust predictions, however, it is always desirable for a solution that allows the socio-temporal interactions to be parameterized and learned as completely and explicitly as possible. 

To this end, we introduce a structure named \textbf{Socio-Temporal Graphs} (STGs, see Fig.~\ref{fig:teaser}). An STG instance not only considers each agent's past trajectories or how they interact at each time instant, but also explicitly describes how one's past trajectory affects the prediction for another one's future trajectory. Also, in contrast to existing work that lets such information be hard-coded into the model \cite{agentformer_iccv21,chen2021human,trajecpred_cvpr22,pred_cvpr22_4}, our model allows this information to be learned. To provide a unified and continuous representation for data sampled jointly from the spatial and temporal domain, we build our model on a self-evolving latent process \cite{vae_arxiv13,bayes_nips21}, with each latent variable encoding a possible structure of the fully-connected directed acyclic graph (DAG), with the weight of each of its directed edge representing how an agent at any time and location affects another agent at another arbitrary time and location.  

We design an attention-based model, named \textbf{STGformer}, that realizes STG and allows it to be learned through an end-to-end pipeline. In particular, when training the model by maximizing the data likelihood, we introduce two separate modules, one representing the prior distribution of the latent sequence and the other representing the posterior of the distribution conditioned on observations. By minimizing the corresponding KL divergence between these two distributions, the latent prior can be learned recursively, so the shape of the distribution is effectively regularized without explicit parametric restrictions.


Our solution achieves state-of-the-art results on Stanford Drone and ETH/UCY trajectory prediction benchmarks with respect to the trajectory-only approaches. Interestingly, our analysis shows that a person's past trajectory is critical for predicting another person's future path, and explicitly utilizing this fact can yield a noticeable performance gain. Furthermore, our experiment indicates that this information exhibits strong spatial-temporal localities.


\bfsection{Contributions}
(1) We introduce the STG structure and formulate multi-agent trajectory forecasting based on learning it with a latent sequential generative model.
(2) We design an attention-based model named STGformer that allows STG to be effectively learned in an end-to-end fashion. 
(3) 
We provide analysis to identify the efficacy of learning STG regarding performance gain.

\section{Related Work}\label{sec:relate}

\bfsection{Multi-agent trajectory prediction}
Current trajectory forecasting approaches exploit a variety of deep neural network models~\cite{bptt_2016}. For instance, some methods adopt a set of LSTMs to characterize the movement of pedestrians~\cite{lstm_cvpr16,gan_cvpr18}. By contrast, NSP-SFM~\cite{pred_eccv22_2} formulates the trajectory forecasting with neural differential equations~\cite{ode_nips19}. PECNet~\cite{trajecpred_4_eccv20} infers the endpoints to assist in trajectory forecasting using a Conditional Variational AutoEncoder (CVAE)~\cite{bptt_2016}. LB-EBM~\cite{trajecpred_cvpr21} generates predictions with a latent energy-based model~\cite{energy_nips19}, while MID~\cite{trajecpred_cvpr22} uses diffusion processes~\cite{diffusion_nips19}.  

Other approaches rely on graphs. Among these, some harness a graph neural network to associate all agents in a fully-connected fashion~\cite{trajecpred_iccv19_1,pred_traj_nips19,trajecpred_eccv20}, while others~\cite{trajecpred_eccv20_1,agentformer_iccv21, pred_cvpr22_2,pred_cvpr22_4,pred_eccv22_3} replace the graph neural network with attention mechanisms~\cite{attn_nips17}. CausalHTP~\cite{chen2021human} and CausalMR~\cite{pred_cvpr22_3} analyze the causal effects with the aid of a pre-defined causal graph~\cite{causal_16} that models the effects from the confounding environmental elements to the agents. 

Most of these studies share the idea of implicitly assuming a fully-connected underlying structure to consider the interactions among the agents per time step. However, the connectivity among agents might be falsely addressed. For example, one agent might be too far from another agent, so considering the motion of one does not help predict the motion of the other. Learning these socio-temporal dependencies and pruning irrelevant edges might reduce overfitting and improve predictions. 



\bfsection{Modeling socio-temporal interactions}
Recent studies have attempted to capture socio-temporal correlations in the context of pedestrian trajectory forecasting. Some studies model socio-temporal interactions implicitly. Among them, a subset first assign a network to each person to obtain temporal cues, and subsequently leverage another network for social cues~\cite{lstm_cvpr16,gan_cvpr18,trajecpred_4_eccv20,trajecpred_eccv20_1,pred_cvpr21_4,socialnce_cvpr21,pred_eccv22_2,pred_traj_nips19,chen2021human, tip09, tip16, tip20, tip21}. by contrast, some recent approaches~\cite{agentformer_iccv21,pred_eccv22_1} tie the social and temporal correlations jointly due to their co-occurrence.

New studies introduced learned time-invariant structures to tackle different tasks, such as physical motion modeling, or to infer the bioinformatic signaling~\cite{causal_nips20,causalvae_cvpr21,bayes_nips21}.
DA-Former~\cite{attn_icml22_1} leverages a greedy strategy to search for an optimal graph for the machine translation problem. 
Following these works, a set of recent techniques explicitly learn a static graph~\cite{pred_iccv21_1,pred_cvpr22_5}. Nevertheless, more than the static representation might be required to characterize the time-varying socio-temporal correlations. EvolveGraph~\cite{pred_traj_nips20} partially addresses this issue with a learned dynamic local social structure. Our work takes inspiration from the aforementioned recent advances, introducing a novel approach to learning time-varying STGs for human trajectory forecasting.

\bfsection{Transformer-based trajectory prediction}
With the success of Transformer models in both fields of natural language processing (NLP) and computer vision, many methods~\cite{yu2020spatio,giuliari2021transformer,agentformer_iccv21,trajecpred_cvpr22,li2022graph,yin2021multimodal} employ the Transformer to build trajectory prediction systems. 
Despite the strong representation ability of the attention model to learn the relations, the Transformer-based framework implicitly builds the fully-connected graph of all agents. This paper shows that we can explicitly learn STG, which reduces ambiguity and significantly improves trajectory prediction. 

\section{Method}
Let $x_i^{t}$ denote the position of agent $i$ at time $t$, and $\bfx^{t}=\{x_i^{t}\}_i$ the positions of all agents at time $t$.  
The goal is to predict $\bfx^{t}$ from the collective past observations $\bfx^{0:t-1}$. 

\begin{figure}[t]
\centering
\includegraphics[width=\linewidth]{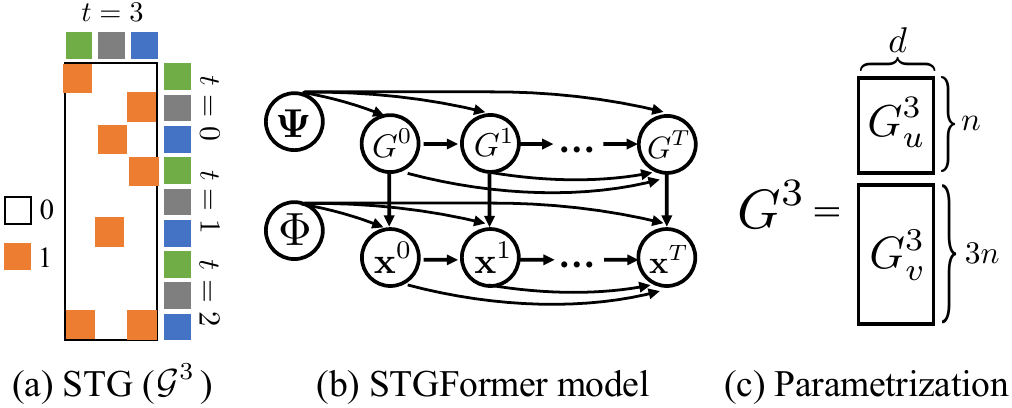}
\vspace{-0.6cm}
\caption{\small
Socio-temporal graph (STG). (a) STG definition as the adjacency matrix between agents at the current step and all previous steps. $\mathcal{G}^3$ means at $t=3$ all agents establish a possible connection with agents over $t \leq 3$. (b) Autoregressive model for trajectory prediction, where $G^t$ is the embedding from an evolving sequence that constructs expanding $\mathcal{G}^3$. (c) $G^3$ has two components, $G_u^3$ and $G_v^3$, which produce $\mathcal{G}^3$ through bilinear product.
}
\label{fig:model}
\vspace{-0.2cm}
\end{figure}

\subsection{Problem Formulation}
\label{subsec:framework}

We define the socio-temporal graph (STG) as a binary adjacency matrix to represent how current observations are related to the past in the form of the directed acyclic graph (DAG). For example, at time $t=3$, directed edges with value $1$ in binary adjacency matrix $\mathcal{G}^3$ (Fig.~\ref{fig:model}\red{a}) indicate a possible one-sided interaction between the agent represented by the source node and the agent represented by the destination node. Such interactions can span over time time $\tau=\{0:t-1\}$ to $t$ to indicate the fact that one's past behavior can influence another's decision in the future. In other words, our hypothesis is that each observation $\bfx^{t}$ is generated not only by  $\bfx^{0:t-1}$, but also the underlying socio-temporal graph, $\mathcal{G}^{t}$.

To learn the dynamics of $\mathcal{G}^{t}$ over time from past observations/training data as opposed to a static one \cite{agentformer_iccv21}, as illustrated in Fig. \ref{fig:model}\red{b}, our model assumes that the probability of the present observation at time $t$ can be decomposed using DAG embedding $G^{t}$ as follows: 
{\small
\begin{align}\label{eq:overall}    
\itp(\bfx^{t}, G^t|\bfx^{0:t-1},G^{0:t-1})=\itp_\Phi(\bfx^{t}|\bfx^{0:t-1},G^{t})\itp_\Psi(G^{t}|G^{0:t-1})
\end{align}}

\noindent Specifically, inspired by the bilinear model for graph generation \cite{arxiv16_vga,bayes_nips21},
$\mathcal{G}^{t}$ is parameterized by $G^{t}$ as:
\begin{align}
    G^{t}=[G_u^{t}; G_v^{t}],\HS\HS\HS\mathcal{G}^t=\left(G_u^{t}(G_v^{t})^T>0\right).
\end{align}
\noindent In a scene of $n$ agents, two embedding matrices $G_u^{t}\in \mathbb{R}^{n\times d}$ and $G_v^{t}\in \mathbb{R}^{nt\times d}$ with $d$ denoting the dimension of the embedded feature (Fig. \ref{fig:model}\red{c}). Through inner product between $G_v^{t}$ and $G_u^{t}$, we have the binary adjacency matrix $\mathcal{G}^{t}$ of the size $n\times nt$, indicating an equal number of pair-wise interactions over both and time-space that our model needs to learn.

\bfsection{STG dynamic module}
We model the STG prior model with a multivariate Gaussian model
\begin{align}\label{eq:overall}    
\itp_\Psi(G^{t}|G^{0:t-1}) = \mathcal{N}(G^{t}; \mu_\Psi(G^{0:t-1}),\textit{I}),
\end{align}which is implemented by one Transformer decoder block~\cite{gpt2} followed by one MLP layer:
\begin{align}\label{eq:G_1}
    \mu_\Psi = \text{MLP}_\textit{G}(\text{TDec}(G^{0:t-1})).
\end{align}

\bfsection{STG-aware trajectory module}
We first introduce the STG-aware attention block and then incorporate it into the trajectory module.

$\mathcal{G}^t$ is obtained using Eqn.~\ref{eq:G_1} to guide the evaluation of self-attention \cite{attn_nips17,gpt2,agentformer_iccv21} as follows:
\begin{align}\label{eq:attn_1}
    \text{STG-aware-attn}^{t} &= \text{softmax}\big(\frac{\mathcal{G}^t\odot(\textup{Q}^t(\textup{K}^t)^T)}{ \sqrt{\textit{n}\times\text{t}}}\big)\textup{V}^t\\
    \label{eq:attn_2}
    \textup{Q}^{t} = \textit{w}_\textup{Q}\bfx^{t-1},
    &\HS\HS\textup{K}^t = \textit{w}_\textup{K}\bfx^{0:t-1},
    \HS\HS\textup{V}^t = \textit{w}_\textup{V}\bfx^{0:t-1},
\end{align}where $\textit{w}_\textup{Q}$, $\textit{w}_\textup{K}$, $\textit{w}_\textup{V}$ are the corresponding weights of the attention heads. $\odot$ calculates the Hadamard product. The attention takes queries $\text{Q}^t$, keys $\text{K}^t$ and values $\text{V}^t$. $\text{K}^t$ and $\text{V}^t$ encode the entire historical trajectories $\bfx^{0:t-1}$. Their dimensions scale with $t$. 
Eqn.~\ref{eq:attn_1} is interoperable with multi-head attention. In addition, STGFormer can mask out the uncorrelated agents, i,e., we assign $\mathcal{G}^{t}_{ij}\odot(\textup{Q}_{i}^t(\textup{K}_{j}^t)^T)=-\infty$ between any query $\textup{Q}_{i}$ of agent $i$ and any key $\textup{K}_{j}$ of agent j, if $\mathcal{G}^{t}_{ij}=0$.

Similarly, to construct $\itp_\Phi(\bfx^{t}|\bfx^{0:t-1},G^{t})$, we have
\begin{align}\label{eq:x}
    \itp_\Phi(\bfx^{t}|\bfx^{0:t-1},G^{t}) = \mathcal{N}(\bfx^{t};\mu_\Phi(\bfx^{0:t-1},G^{t}), I).
\end{align}
\noindent with mean $\mu^{t}$ and an identity matrix of a Multivariate Gaussian:
\begin{align}\label{eq:x_1}
    \mu_\Phi = \text{MLP}_\textit{x}(\text{STG-aware-attn}^{t})
\end{align}
We use another Transformer decoder, as well as an MLP to instantiate $\mu_\Phi$. The prediction of $\bfx^{t}$ is then sampled according to the output of Eqn.~\ref{eq:x_1}, using our proposed STG-aware attention (STG-aware-attn).



\begin{figure*}[t]
\centering
\includegraphics[width=1\linewidth]{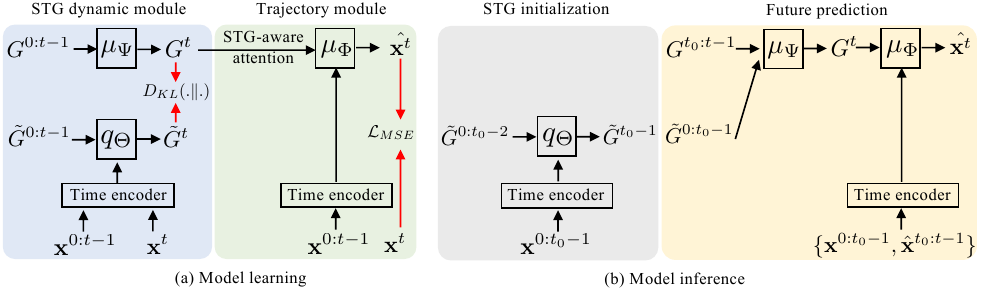}
\vspace{-0.6cm}
\caption{
Learning and inference for STGFormer model. 
(a) During training, we minimize the STG prediction from the prior and posterior modules and the reconstruction from the trajectory module. (b) During inference, we first initialize STGs with the posterior module and then predict future trajectories with the prior and trajectory modules.}
\label{fig:stgformer}
\vspace{-0.4cm}
\end{figure*}
\subsection{Learning}
\bfsection{Variational Inference} The goal of training is to determine $\Psi$ and $\Phi$ that maximize the log likelihood of the observed data $\itp(x)$. As obtaining the exact solution is intractable, we instead seek to evaluate its evidence lower bound (ELBO) with a posterior model for STG $\itq_\Theta(\Tilde{G}^{t}|\Tilde{G}^{0:t-1}, \bfx^{0:t})$.
We denote the STG samples from the posterior model as $\tilde{G}^t$, where those from the prior model as $G^t$.
{\small 
\begin{align}\label{eq:elbo}
    &\log\itp(\bfx^{0:T}) \geq L(\Theta,\Phi,\Psi) =  
    \mathbb{E}_{G^{0:T}\sim q_{\Theta}}\log\frac{\itp(\bfx^{0:T}, G^{0:T})}{\itq_{\Theta}(\Tilde{G}^{0:T}|\bfx^{0:T})}\nonumber\\     
    &= \mathbb{E}_{G^{0:T}\sim q_{\Theta}} \log\prod\limits_{t=0}^{T}\left(\frac{\itp_\Phi(\bfx^{t}|\bfx^{0:t-1},G^{t})\itp_\Psi(G^{t}|G^{0:t-1})}{\itq_{\Theta}(\Tilde{G}^{t}|\Tilde{G}^{0:t-1}, \bfx^{0:t})}\right)\nonumber\\
    &=\sum\limits_{t=0}^{T}\mathbb{E}_{G^{0:T}\sim q_{\Theta}} \log\left(\frac{\itp_\Phi(\bfx^{t}|\bfx^{0:t-1},G^{t})\itp_\Psi(G^{t}|G^{0:t-1})}{\itq_{\Theta}(\Tilde{G}^{t}|\Tilde{G}^{0:t-1}, \bfx^{0:t})}\right) \nonumber\\
    & =\sum\limits_{t=0}^{T} \left[\mathbb{E}_{G^{t}\sim q_\Theta} \log\itp_\Phi(\bfx^t|\bfx^{0:t-1},G^t)      \right. \nonumber\\         
    &\left.\HS\HS\HS\HS\HS\HS\HS\HS\HS\HS\HS\HS-D_{KL}\left(q_{\Theta}(\Tilde{G}^{t}|\Tilde{G}^{0:t-1},\bfx^{0:t})\|\itp_\Psi(G^{t}|G^{0:t-1})\right)\right]
\end{align}}where $(.)^{0:t-1}$ is an empty set when $t$=0.

\bfsection{Training objective} 
First, due to the normality of $p_\Phi$, the first term in Eqn.~\ref{eq:x} reduces to the mean square error loss
\begin{align}L_{MSE} = \sum\limits_{t=0}^{T} \mathbb{E}_{G^{t}\sim q_\Theta}(||\bfx^t - \mu_{\Phi}(\bfx^{0:t-1},G^t)||^2),
\end{align}where the expectation is evaluated numerically.

Then, we assume the posterior also takes a form of conditional Multivariate Gaussian:  
\begin{align}
    \itq_\Theta(\Tilde{G}^{t}|\Tilde{G}^{0:t-1}, \bfx^{0:t})= \mathcal{N}(\mu_{\Theta}(\bfx^{0:t}, \Tilde{G}^{0:t-1}),\textit{I}),
\end{align}where $\mu(\cdot; \Theta)$ is produced by a separate Transformer decoder \cite{gpt2} parameterized by $\Theta$.  
Due to the nomality of both distributions, we define the KL loss $L_{KL}$ involving the learnable parameters in the KL-divergence term analytically 
\begin{align}
L_{KL} 
& = \sum\limits_{t=0}^T \|\mu_{\Psi}(G^t|G^{0:t-1})-\mu_{\Theta}(\Tilde{G}^{t}|\bfx^{0:t},\Tilde{G}^{0:t-1})\|^2.
\end{align}
This way, ${G}^t$ becomes data-dependent with its samples directly forming a loss function for differentiable learning. In other words, our model allows learnable socio-temporal interactions as they are encoded by edges in a corresponding adjacency matrix whose distribution is determined by training data. 


To further regularize the distribution of the resulting structure $\mathcal{G}^t$, we also apply a $L_0$-norm penalty. 
In conclusion, we have the following training objective function
\begin{align}\label{eq:loss_total}
L = L_{MSE}+L_{KL}+\zeta\|\mathcal{G}^{0:T}\|_0.
\end{align}




\subsection{Inference}
During the initialization, given the known trajectory $\bfx^{0:t_0-1}$, we can use the posterior model to generate $\Tilde{G}^{0:t_0-1}$. We then feed $\Tilde{G}^{0:t_0-1}$ to the prior module to estimate $G^{t_0}$ from Eq.~\ref{eq:overall} and \ref{eq:G_1}. This design exploits the historical data to learn the reasonable $G^{t}$ for predictions. To generate plausible future trajectories predictions, we sample from $\itp_\Phi(\bfx^{t}|\bfx^{0:t-1},G^{t})$, following Eqn.~\ref{eq:x} and \ref{eq:x_1}.

Again, we expect the distribution of our prediction maximizes the ELBO by minimizing the KL divergence in Eqn.~\ref{eq:elbo} as the underlying hypothesis is that predictions over new observations and old samples used for training are drawn from a similar distribution being captured by our learned $G^t$.
  


\section{Implementation Details}

\subsection{STG-aware trajectory module $\itp_\Phi(\bfx^{t} |G^{t}, \bfx^{0:t-1})$}

\begin{figure*}[t]
\centering
\includegraphics[width=\linewidth]{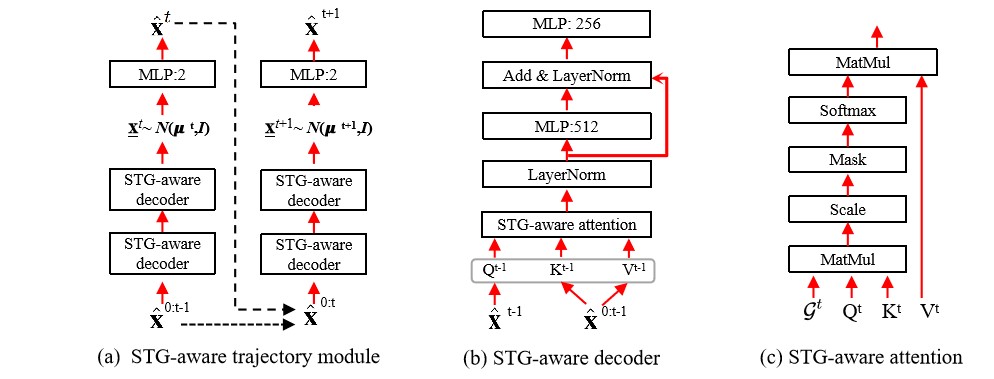}
\caption{Architecture details for our STG-aware trajectory module $\itp_{\Phi}(\bfx^{t}|\bfx^{0:t-1},G^{0:t})$. (a) The outcomes from the first STG-aware decoder serve to generate the queries, keys, and values for the second layer. We treat the output from the MLP that perceives $\underline{x}^t$ as final predictions. (b) We replace the masked attention mechanism in the standard transformer decoder with our STG-aware attention. (c) STG-aware attention in addition takes the STGs $G^{0:t}$ as input.}  
\label{fig:traj_model}
\end{figure*}

Our method leverages masked multi-head attention \cite{attn_nips17}. Specifically, the time encoder with sequential representation produces a sequence of path features $\bfx^t = \{x^t_0, x^t_1... x^t_{N-1}\}$ for all N pedestrians in the scene~\cite{attn_iccv21}. The STG-aware Attention has been defined by Eq.~\red{5,6} in the main paper.
Fig.~\ref{fig:traj_model}\red{b} visualizes how STG-aware decoder proceeds after receiving STG-aware Attention (Fig.~\ref{fig:traj_model}\red{c}). Specifically, we have:
 \begin{align}\label{eq:tm}
    & u^{t} = \text{LayerNorm}(\text{STG-aware-attn}^{t})\nonumber\\ 
    & 
    a^{t} = \text{LayerNorm}\Big(u^{t} + \text{MLP}\big(\text{ReLU}(\text{MLP}(u^{t}))\big)\Big)\nonumber\\
    & 
    \mu^{t} = \text{MLP}(a^{t})
\end{align}

Further, Fig.~\ref{fig:traj_model}\red{a} showcases that the prediction of $\hat{x}^{t}$ is obtained by 
\begin{align}\label{eq:x}
  & \underline{x}^t \sim \mathcal{N}(\mu^{t}, \textit{I})\nonumber\\
    & 
\hat{x}^{t} = \text{MLP}(\underline{x}^t)
\end{align}

\begin{figure*}[t]
\centering
\includegraphics[width=0.8\linewidth]{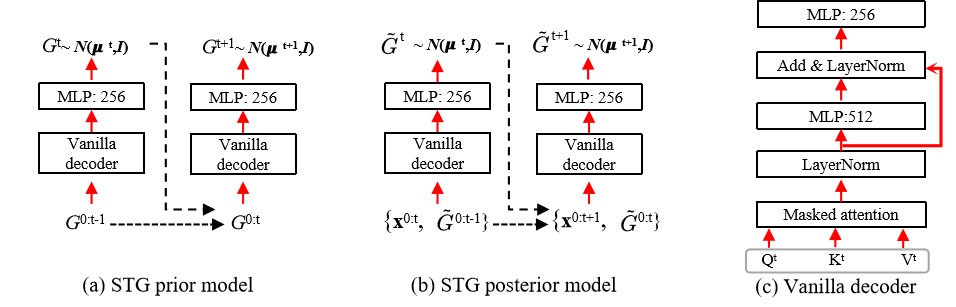}
\caption{Architecture details for our STG prior model $\itp_\Psi(G^{t}|G^{0:t-1})$ and STG posterior model $\itq_\Theta(\Tilde{G}^{t}|\Tilde{G}^{0:t-1}, \bfx^{0:t})$. (a) The transition of the sequence $G^t$ of our STG prior model. (b) Our STG posterior model perceives both $\bfx^{0:t}$ and $\tilde{G}^{0:t-1}$ to generate $\tilde{G}^t$. (c) We use the vanilla transformer decoder \cite{gpt2} to implement both the STG prior model and STG posterior model.}
\label{fig:prior}
\end{figure*}

\subsection{ STG prior model $\itp_\Psi(G^{t}|G^{0:t-1})$}
As indicated in Figure \ref{fig:prior}\red{a}, at each time instant $t$, $G^t$ is directly sampled from a Gaussian distribution whose mean and variance are regressed from $G^{0:t-1}$. More specifically, at $t-1$, to allow $G^{t-1}$ and $G^{t}$ to have the same dimension, we pad $G^{t-1}$ with zeros to generate the queries, $G^{0:t-1}$ models the corresponding keys and values in the vanilla transformer decoder \cite{gpt2}. In practice, we assume $G^0\sim N(0,1)$ for training. We formulate the generative process of $G^t$ in the following:
\begin{align}\label{eq:lm}
    & \textup{Q}^{t} = \textit{w}_\textup{Q}^{L}[\text{Concat}(G^{t-1};0)],
    \HS\HS\textup{K}^t = \textit{w}_\textup{K}^{L}(G^{0:t-1}),
    \HS\HS\textup{V}^t = \textit{w}_\textup{V}^{L}(G^{0:t-1})\nonumber\\ 
    & \text{A}^{t} = \text{softmax}\big(\frac{\textup{Q}^t(\textup{K}^t)^T}{\sqrt{\textit{n}\times\sum\limits_{t}t}}\big)\textup{V}^t\nonumber\\
    & u^{t} = \text{LayerNorm}(\text{A}^{t})\nonumber\\ 
    & 
    a^{t} = \text{LayerNorm}\Big(u^{t} + \text{MLP}\big(\text{ReLU}(\text{MLP}(u^{t}))\big)\Big)\nonumber\\
    & 
    \mu^{t} = \text{MLP}(a^{t}) \nonumber\\
    & G^t \sim \mathcal{N}(\mu^{t}, \textit{I}) 
\end{align}

\subsection{STG posterior model $\itq_\Theta(\Tilde{G}^{t}|\Tilde{G}^{0:t-1}, \bfx^{0:t})$}
Figure \ref{fig:prior}\red{b} shows how samples of $\tilde{G}^t$ are drawn from $\itq_\Theta(\Tilde{G}^{t}|\Tilde{G}^{0:t-1}, \bfx^{0:t})$.
In particular, $\tilde{G}^0$ is trained by fitting the posterior  $\tilde{G}^0\sim\itq_\Theta(\Tilde{G}^0 | \bfx^0)$ to the prior $G^0\sim N(0,1)$. From this point onward ($t\geq 1$), 
$\bfx^{0:t}$ are treated as queries and $\tilde{G}^{0:t-1}$ are used to generate keys and values. The STG posterior model can be written with the following:
\begin{align}\label{eq:qm}
    & \textup{Q}^{t} = \textit{w}_\textup{Q}^{P}(\bfx^{0:t}),
    \HS\HS\textup{K}^t = \textit{w}_\textup{K}^{P}(\tilde{G}^{0:t-1}),
    \HS\HS\textup{V}^t = \textit{w}_\textup{V}^{P}(\tilde{G}^{0:t-1}) \nonumber\\   
    & \text{A}^{t} = \text{softmax}\big(\frac{\textup{Q}^t(\textup{K}^t)^T}{\sqrt{\textit{n}\times\sum\limits_{t}t}}\big)\textup{V}^t\nonumber\\
    & u^{t} = \text{LayerNorm}(\text{A}^{t})\nonumber\\ 
    & 
    a^{t} = \text{LayerNorm}\Big(u^{t} + \text{MLP}\big(\text{ReLU}(\text{MLP}(u^{t}))\big)\Big)\nonumber\\
    & 
    \mu^{t} = \text{MLP}(a^{t}) \nonumber\\
    & \Tilde{G}^t \sim \mathcal{N}(\mu^{t}, \textit{I})
\end{align}

\bfsection{Network training}
We clip the maximum value of the KL term in Eqn.~\ref{eq:elbo} down to 2. In our experiments, AdamW optimizer \cite{adamw_iclr19}, and cosine annealing are employed with the learning rate initialized at 10$^{-3}$ with a momentum of 0.9 and weight decay of 10$^{-2}$. Our models are implemented using PyTorch, and experiments are conducted on four Nvidia GeForce 2080Ti graphics cards.

\begin{table}[t]
\begin{center} 
\caption{
Results on the SDD dataset with best-of-20 ADE/FDE scores. 
For fair comparison, We rank methods with trajectory input only (T) 1st in bold and 2nd underlined. Also we showcase the results from those with extra image input (T+I).}
\vspace{-0.2cm}
{
\begin{tabular}{l|c|c}
\hline
&  Input & ADE/FDE $\downarrow$ \\ \hline
Y-Net~ \cite{y-net_iccv21}
& T + I & 7.85/11.85\\
V$^2$-Net \cite{pred_eccv22_1} & T + I & 7.12/11.39\\
NSP-SFM \cite{pred_eccv22_2} & T + I & 6.52/10.61\\
Muse-VAE~ \cite{muse_vae}
& T + I & 6.36/11.10\\
\hline
Social-GAN \cite{gan_cvpr18}
& T & 27.23/41.44
\\
EvolveGraph \cite{pred_traj_nips20}
& T & 13.90/22.90
\\
GroupNet
\cite{pred_cvpr22_5} 
& T &9.31/16.11\\
GP-Graph\cite{pred_eccv22_4}
& T & 9.10/13.80\\
LB-EBM~
\cite{trajecpred_cvpr21}
& T & 8.87/15.61\\
PCCSNET~
\cite{trajecpred_iccv21}
& T & 8.62/16.16\\
NPSN \cite{pred_cvpr22_1} & T & 8.56/\underline{11.85}\\
MemoNet \cite{pred_cvpr22_4} & T & 8.56/12.66\\
AgentFormer
\cite{agentformer_iccv21} 
& T & 8.35/13.03\\
Expert \cite{expert} & T & 7.69/14.38\\
MID~
\cite{trajecpred_cvpr22} 
& T & \underline{7.61}/14.30\\
\hline
\textbf{\model~(Ours)}
& T &  \bf{7.35}/\textbf{11.39} \\
\hline
\end{tabular}
}
\label{tab:sdd}
\end{center}
\vspace{-0.6cm}
\end{table}

\begin{table*}[t]
\begin{center} 
\caption{The benchmark Best-of-20 ADE/FDE results ($\downarrow$) on the ETH/UCY datasets. For fair comparison, We rank methods with trajectory input only (T) 1st in bold and 2nd underlined. Also we showcase the results from those with extra image input (T+I). {\color{red}$^\dagger$} Up-to-date results from the official implementations are worse than the original ones due to an evaluation bug.
}
{
\begin{tabular}{l|c|ccccc|c}
\hline
&  { Input} &
{ ETH}  & { Hotel} &  { Univ.}&{ Zara1}&{ Zara2}&{ AVG}  \\ \hline
Social-BiGAT~ 
\cite{pred_traj_nips19}
& T + I  & 0.69/1.29  & 0.49/1.01 & 0.30/0.62  &   0.36/0.75 & 0.55/1.32 & 0.48/1.00 \\
Y-Net~ \cite{y-net_iccv21}
& T + I & 0.28/0.33 & 0.10/0.14 & 0.24/0.41 & 0.17/0.27 & 0.13/0.22 & 0.18/0.27 \\
V$^2$-Net \cite{pred_eccv22_1} & T + I & 0.23/0.37 & 0.11/0.16 & 0.21/0.35  & 0.19/0.30 & 0.14/0.24 & 0.18/0.28 \\
NSP-SFM \cite{pred_eccv22_2} & T + I & 0.25/0.24 & 0.09/0.13 & 0.21/0.38  & 0.16/0.27 & 0.12/0.20 & 0.17/0.24 \\
\hline
Social-GAN \cite{gan_cvpr18}
& T & 0.81/1.52 & 0.72/1.61 & 0.60/1.26 & 0.34/0.69 & 0.42/0.84 & 0.58/1.18 
\\
CausalHTP~
\cite{chen2021human}
& T & 0.60/0.98  & 0.30/0.54 & 0.32/0.64  &   0.28/0.58 & 0.52/1.10  & 0.40/0.77 \\ 
Trajectron++~\red{$^\dagger$}
\cite{trajecpred_eccv20}
& T & 0.67/1.18  & 0.18/0.28 & 0.30/0.54  & 0.25/0.41 & 0.18/0.32 & 0.32/0.55\\
SG-Net~\red{$^\dagger$} \cite{sg-net_iros21}
& T & 0.47/0.77  & 0.20/0.38 &   0.33/0.62  & 0.18/0.32 & 0.15/0.28 & 0.27/0.47 \\
STAR~
\cite{trajecpred_eccv20_1} 
& T & 0.36/0.65  & 0.17/0.36 &  0.31/0.62  &   0.26/0.55  &  0.22/0.46  & 0.26/0.53 \\
GroupNet~\red{$^\dagger$}
\cite{pred_cvpr22_5} 
& T &  0.46/0.73  & 0.15/0.25 &  0.26/0.49  &   0.21/0.39  &  0.17/0.33  &  0.25/0.44 \\
GP-Graph\cite{pred_eccv22_4}
& T & 0.43/0.63  & 0.18/0.30 & 0.24/0.42  &  0.17/\underline{0.31} & 0.15 0.29 & 0.23/0.39 \\
AgentFormer~\red{$^\dagger$}
\cite{agentformer_iccv21} 
& T & 0.45/0.75  & 0.14/0.22 & 0.25/0.45  &  0.18/\bf{0.30} & 0.14/\bf 0.24 & 0.23/0.39\\
LB-EBM~
\cite{trajecpred_cvpr21}
& T & 0.30/\bf 0.52 & \underline{0.13}/0.20 & 0.27/0.52 &  0.20/0.37  & 0.15/0.29 & 0.21/0.38\\
PCCSNET~
\cite{trajecpred_iccv21}
& T & \underline{0.28/0.54}  & {\bf 0.11}/0.19  & 0.29/0.60 &  0.21/0.44  & 0.15/0.34 & 0.21/0.42\\
MID~
\cite{trajecpred_cvpr22} 
& T &  0.39/0.66  & \underline{0.13}/0.22 & \underline{0.22}/0.45  &  0.17/\bf 0.30 & \underline{0.13}/0.27 & 0.21/0.38\\
NPSN \cite{pred_cvpr22_1} & T & 0.36/0.59 & 0.16/0.25 & 0.23/\bf 0.39  & 0.18/0.32 & 0.14/\underline{0.25} & 0.21/\underline{0.36}\\
MemoNet \cite{pred_cvpr22_4} & T & 0.40/0.61 & {\bf 0.11}/\underline{0.17} & 0.24/\underline{0.43}  & 0.18/0.32 & 0.14/ \bf 0.24 & 0.21/\bf 0.35\\
Expert \cite{expert} & T & 0.37/0.65 & \bf{0.11/0.15} & {\bf 0.20}/0.44  & {\bf 0.15}/\underline{0.31} & {\bf 0.12}/0.26 & \underline{0.19/0.36}\\
\hline
\textbf{\model~(Ours)}
& T & {\bf0.27}/0.56  & {\bf0.11}/\underline{0.17} & \underline{0.22}/0.45 & \underline{0.16/ 0.31}  & 0.14/\bf 0.24  & \bf 0.18/0.35 \\
\hline
\end{tabular}
}
\label{tab:benchmark}
\end{center}
\end{table*}

\section{Experiments}\label{sec:exp}
We conducted extensive experiments to justify the effectiveness of our approach in terms of trajectory prediction. We also conduct experiments on the information captured by the STG by examining its existence, efficacy, and socio-temporal characteristics. 

\subsection{Experimental Setup}

\bfsection{Datasets} The performance of our approach to trajectory forecasting was evaluated on two widely used datasets: ETH/UCY \cite{social_iccv09,ucy_07} and the Stanford Drone Dataset (SDD) \cite{sdd_eccv16}. 
The ETH/UCY dataset contains five scenes, most of which include more than 700 different pedestrians. The human trajectories are captured in real-world scenes.
The SDD dataset comprises long video sequences for 20 scenes captured using a drone in a top-down view around a university campus. It labels complete trajectories of different categorized moving objects (e.g., pedestrians, bicyclists, and vehicles) from entering the scene until they exit. Both datasets include highly dynamic scenarios with rich socio-temporal correlations, and as such, they are ideal for evaluating the performance of our approach.

Following state-of-the-art methods, \cite{gan_cvpr18,trajecpred_iccv19_1,pred_traj_nips19,agentformer_iccv21,chen2021human}, the trajectories of all datasets are sampled at 0.4 seconds intervals. The first 3.2 seconds (8 frames) are observed for each video, and the next 4.8 seconds (12 frames) are to be predicted. We adopt a leave-one-scene-out approach for our experiments on the ETH/UCY dataset, training and validating our model on videos in 4 scenes and testing on the 5th scene. We repeat this procedure for all five scenes. In addition, we apply the same training and testing procedure for all baseline methods to ensure a fair comparison. 

\bfsection{Evaluation metrics}
We follow the same evaluation metrics adopted by previous work \cite{pred_traj_nips19,trajecpred_eccv20_1,agentformer_iccv21,chen2021human,trajecpred_cvpr22}, including the Best-of-20 average displacement error (ADE) and final displacement error (FDE). In addition, our evaluation is performed for each agent compared to the respective ground truth, where ADE computes the mean square error (MSE) of 20 predictions and the ground truth, while FDE calculates the L2 distance between the final locations of 20 predictions and the ground truth. 

\subsection{Comparing with State-of-the-art Methods}

\bfsection{Methods in comparison}
We categorize methods as Trajectory-Only (T) and Trajectory-and-Image (T+I) methods, as the additional image information may be crucial in certain circumstances yet increases computation cost.
Methods using T+I include Y-Net\cite{y-net_iccv21}, V$^2$-Net \cite{pred_eccv22_1}, NSP-SFM \cite{pred_eccv22_2}, Muse-VAE \cite{muse_vae}, and Social-BiGAT \cite{pred_traj_nips19}. Methods using T-only include GP-Graph \cite{pred_eccv22_4}, MID \cite{trajecpred_cvpr22}, MemoNet \cite{pred_cvpr22_4}, NPSN \cite{pred_cvpr22_1}, GroupNet \cite{pred_cvpr22_5}, Expert \cite{expert}, PCCSNET \cite{trajecpred_iccv21}, LB-EBM \cite{trajecpred_cvpr21}, Trajectron++ \cite{trajecpred_eccv20}, Y-Net\cite{y-net_iccv21}, AgentFormer \cite{agentformer_iccv21}, EvolveGraph \cite{pred_traj_nips20}, Social-GAN \cite{gan_cvpr18}, Expert \cite{expert}, CausalHTP \cite{chen2021human}, STAR \cite{trajecpred_eccv20_1}, Trajectron++ \cite{trajecpred_eccv20}, SG-net\cite{sg-net_iros21}, and Social-GAN \cite{gan_cvpr18}.

\bfsection{Quantitative results on SDD dataset}
Tab.~\ref{tab:sdd} shows that \modelo outperforms other approaches with trajectories-only input consistently in ADE and FDE. Notably, our method improves the ADE by achieving a value of 7.35. In terms of FDE, our approach accomplishes a score of 11.39. These findings tip the balance towards our approach with respect to the other approaches.

\bfsection{Quantitative results on ETH/UCY dataset}
In the second part of our benchmark experiments, our results on ETH/UCY datasets are summarized in Tab.~\ref{tab:benchmark}. 
Experiment results demonstrate that our proposed method achieves the best average 0.18 ADE and 0.35 FDE over the trajectory-only approaches.

\begin{figure}[t]
\centering   
\includegraphics[width=\linewidth]{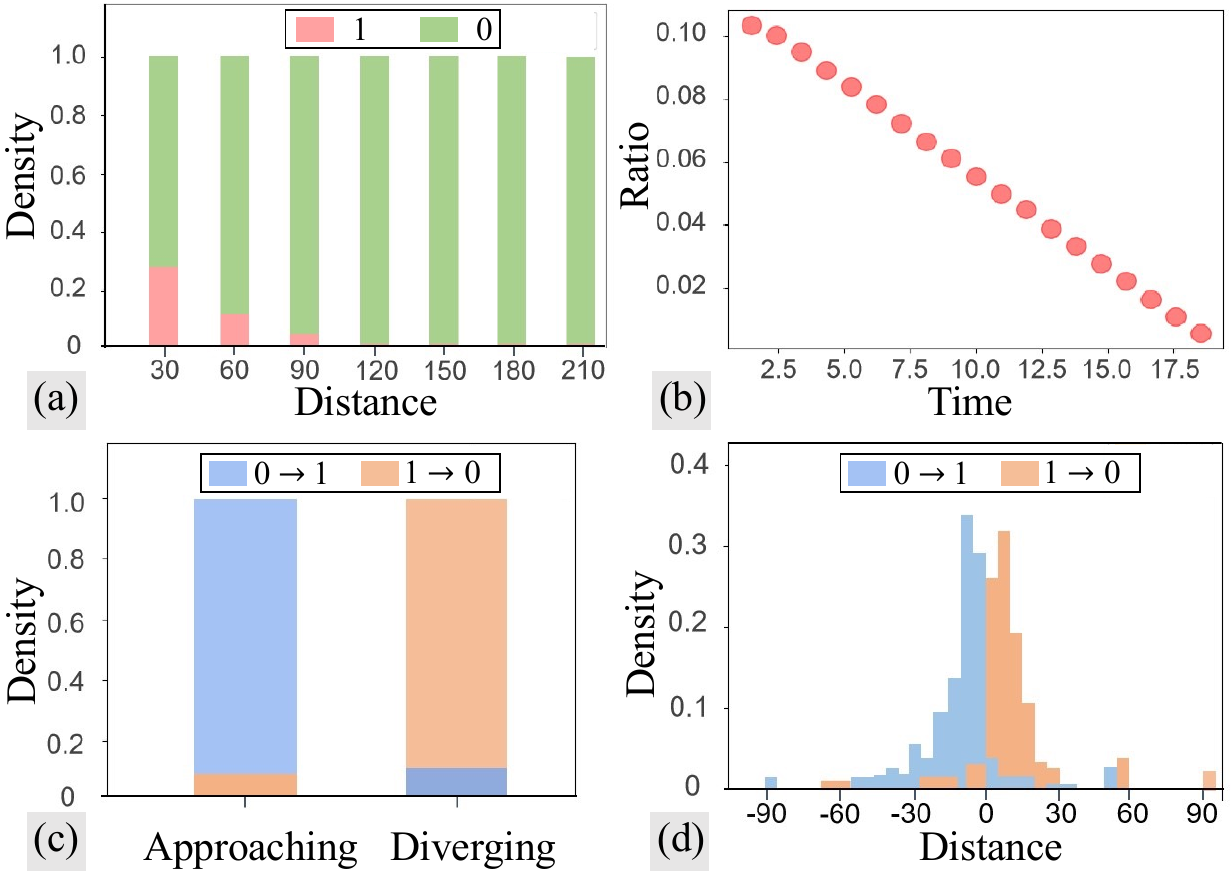}
    \caption{We compute the statistics of the learned edge values and study the socio-temporal notion of the pair-wise agent interactions these edges encode. (a) The sparse distribution of learned non-trivial STG edge values over the distance between the two agents the edges connect. Edges with a learned value of $1$ mostly link two agents no more than of $120$-pixel, meaning the behaviors of most agents outside this range are deemed less essential for trajectory predictions. (b) The distribution of edges with a value $1$ over time. ``Short-term'' edges are more likely to be considered than their ``long-term'' counterparts for trajectory predictions.
    (c) The flip of edge from $0$ to $1$ is highly correlated with the event where two agents  ``approaching" each other, and vice-versa, the event that two agents ``diverging" from each other are captured by the edge value flipping from $1$ to $0$. (d) Symmetry and spatial localization suggest that most events occur when one agent is entering or leaving another agent's $30$-pixel-perimeter.
    }
    \label{fig:stat}
    \vspace{-0.4cm}
\end{figure}

\begin{figure*}[t]
  \centering  
  \text{ETH\HS\HS\HS\HS\HS\HS\HS\HS\HS\HS\HS\HS\HS\HS\HS\HS\HS\HS\HS\HS\HS\HS\HS\HS
Hotel\HS\HS\HS\HS\HS\HS\HS\HS\HS\HS\HS\HS\HS\HS\HS\HS\HS\HS\HS\HS\HS\HS\HS\HS Univ\HS\HS\HS\HS\HS\HS\HS\HS\HS\HS\HS\HS\HS\HS\HS\HS\HS\HS\HS\HS  Zara1 \HS\HS\HS\HS\HS\HS\HS\HS\HS\HS\HS\HS\HS\HS\HS\HS\HS\HS\HS\HS\HS
 Zara2}\\
\vspace{0.2cm}
    \includegraphics[width=0.19\linewidth]{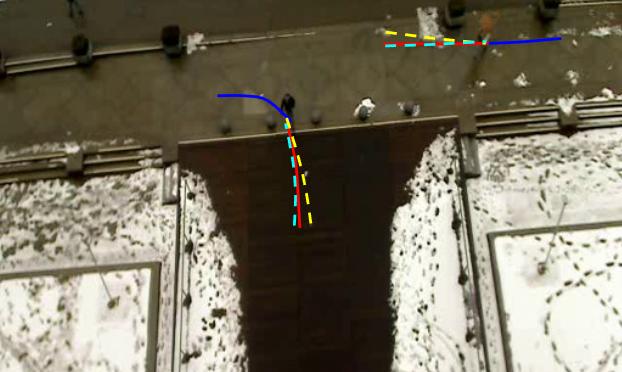}
    \includegraphics[width=0.19\linewidth]{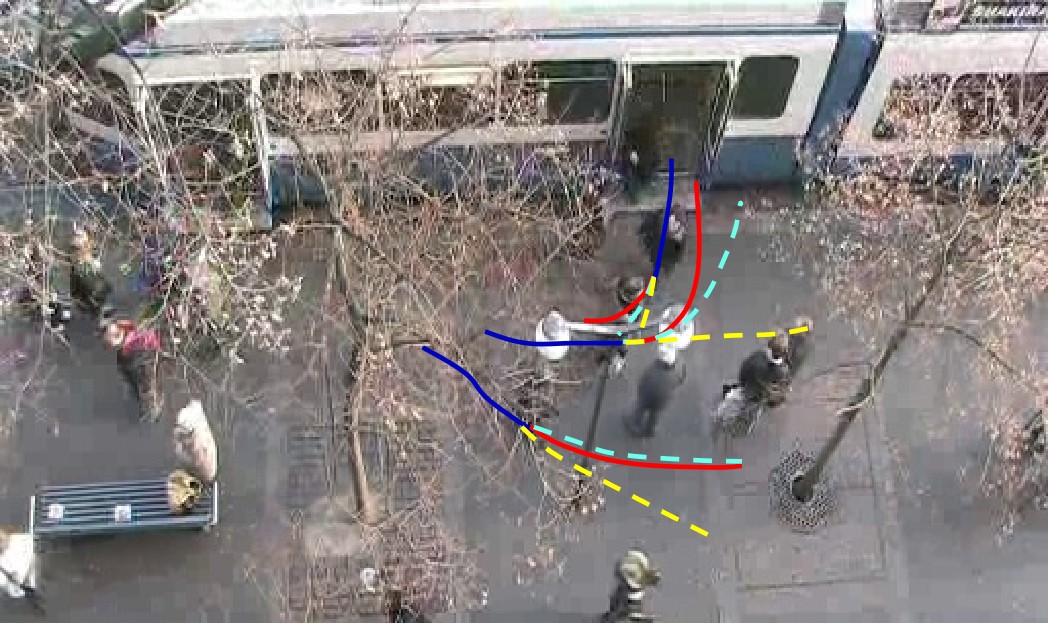}
    \includegraphics[width=0.19\linewidth]{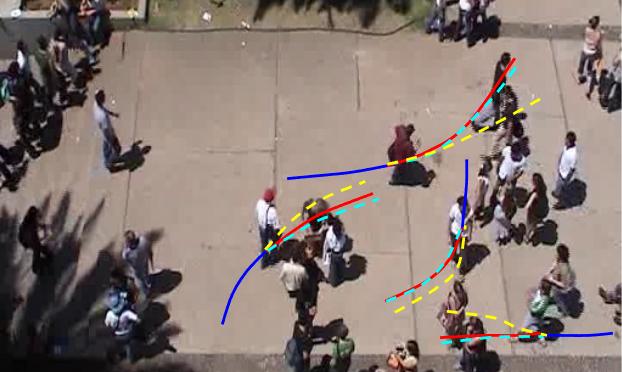}
    \includegraphics[width=0.19\linewidth]{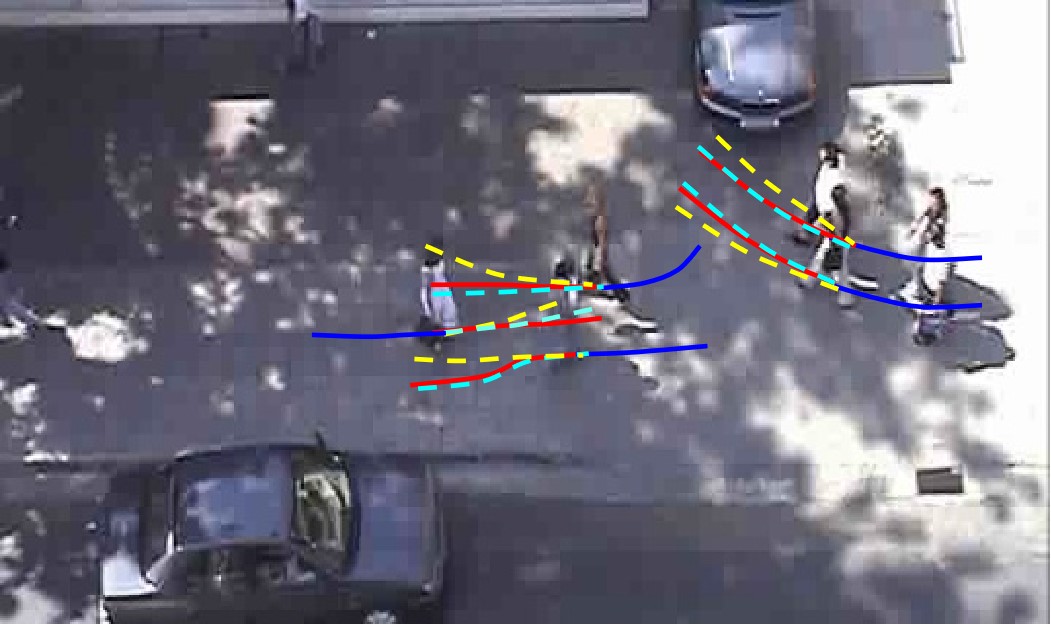}
    \includegraphics[width=0.19\linewidth]{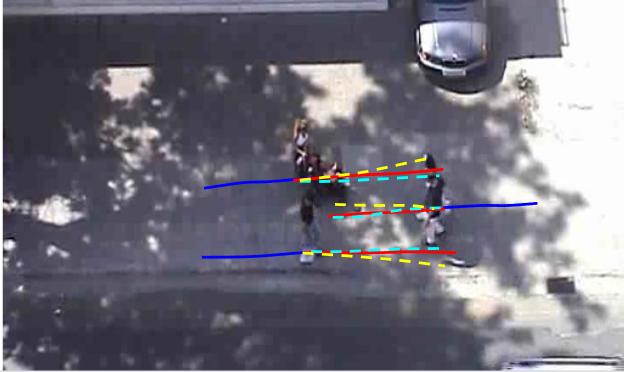}\\
    \vspace{0.2cm}
    \includegraphics[width=0.19\linewidth]{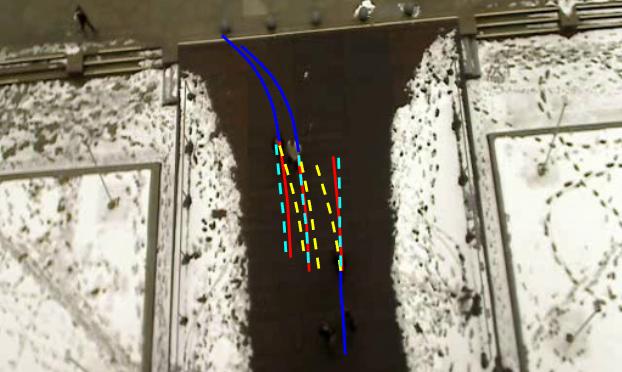}
    \includegraphics[width=0.19\linewidth]{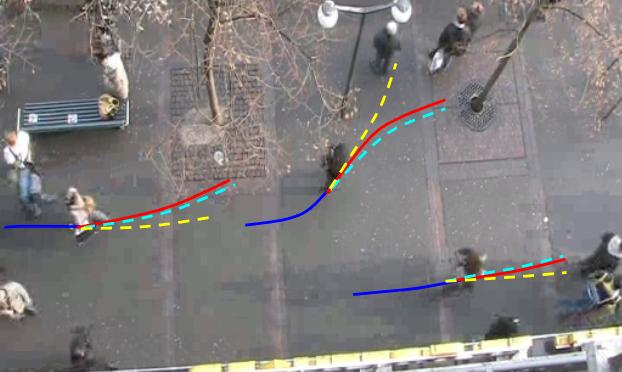}
    \includegraphics[width=0.19\linewidth]{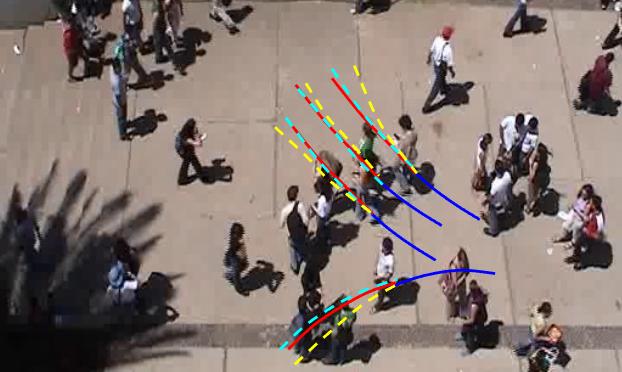}
    \includegraphics[width=0.19\linewidth]{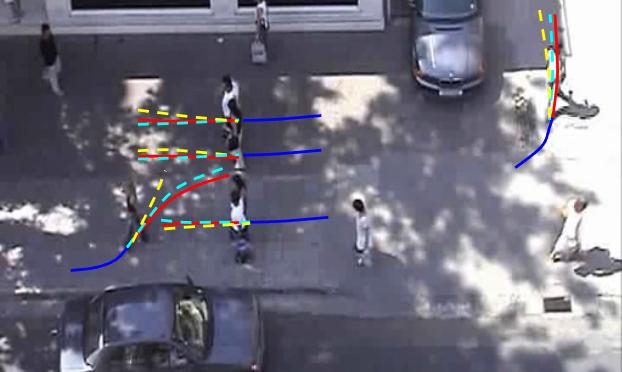}
    \includegraphics[width=0.19\linewidth]{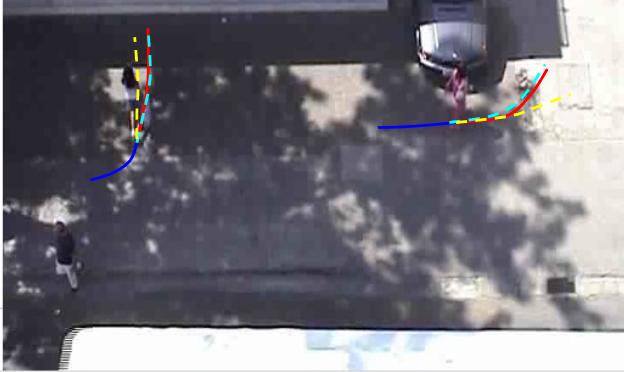} \\   
    \vspace{0.2cm}
    \includegraphics[width=0.7\linewidth]{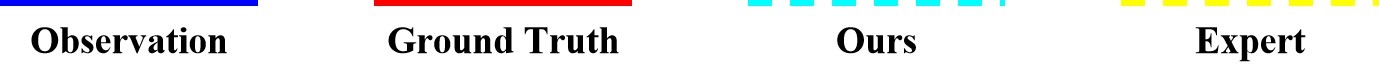} \\
  \caption{Qualitative results on ETH/UCY datasets. Each column overlays predictions for several individual agents. For each person, we show the path history (blue line), ground truth future (red line), and predictions from Expert~\cite{expert} (yellow dashed line) and ours \model~(cyan dashed line). }
  \label{fig:eth}
\end{figure*}

\begin{figure}[t]
  \centering
    \includegraphics[width=0.48\linewidth]{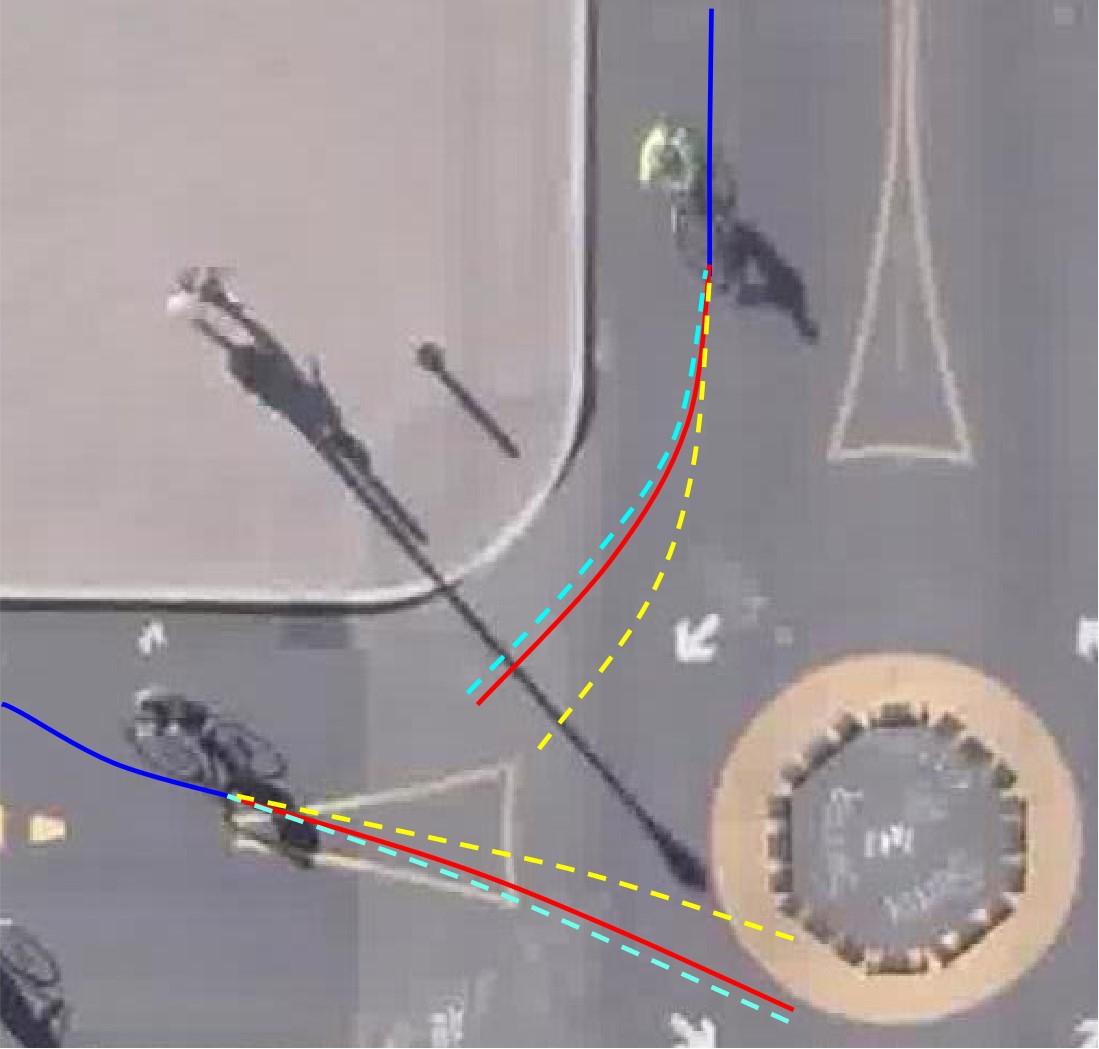}
    \includegraphics[width=0.48\linewidth]{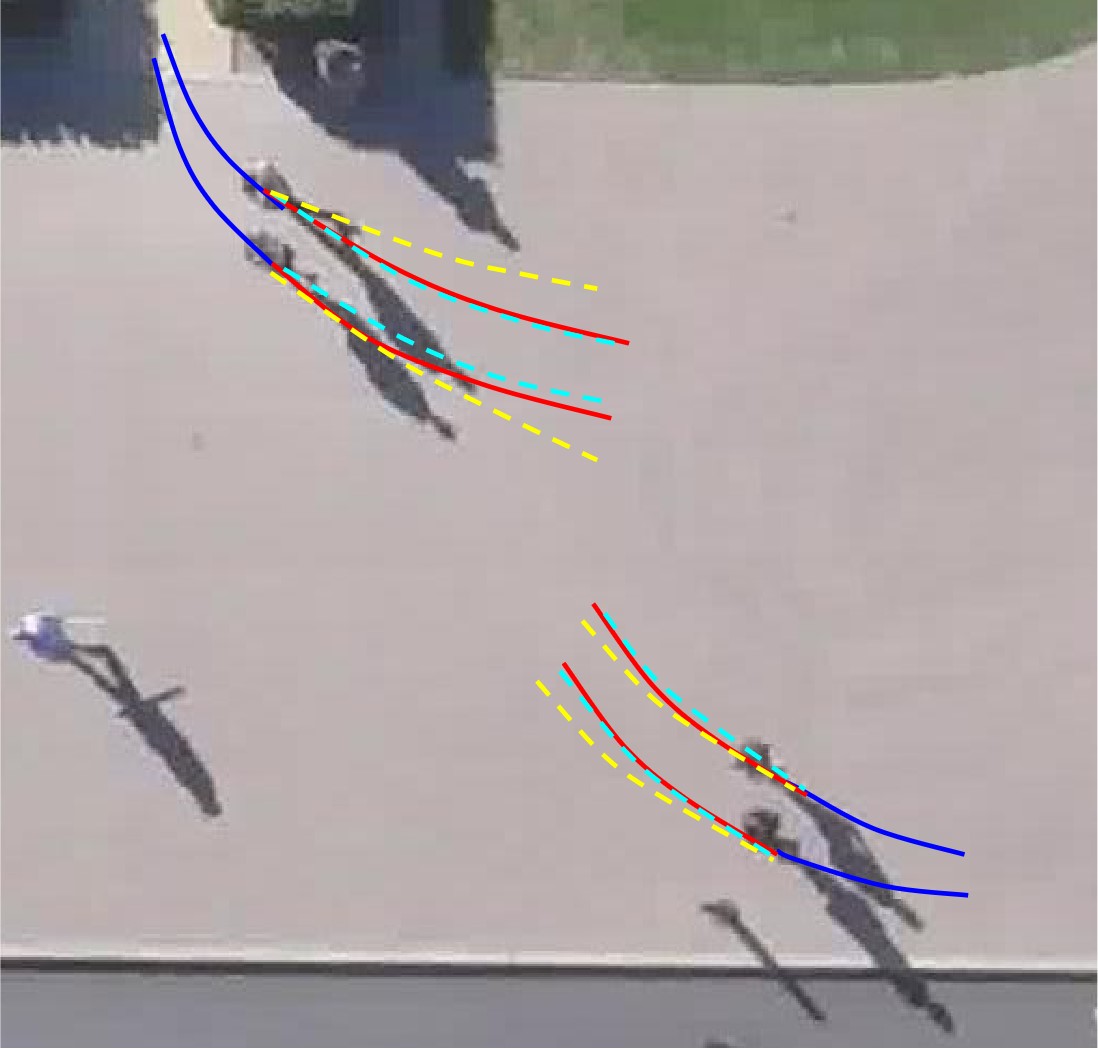}\\
  \vspace{0.1cm}
  \includegraphics[width=\linewidth]{images/label_eth.jpg}\\
\caption{Qualitative results on SDD. Each example overlays the predictions from MID~\cite{trajecpred_cvpr22} (yellow dashed line) and ours \model~(cyan dashed line), historical observations (blue line), and ground truth future (red line)}
\label{fig:sdd}
\end{figure}

\bfsection{Qualitative results}
Fig.~\ref{fig:sdd} and Fig.~\ref{fig:eth} showcase the most-likely prediction from \model~and Expert~\cite{expert} on SDD and ETH/UCY datasets. 
\model~generates more accurate predictions concerning the true future than Expert~\cite{expert}, especially for longer-term predictions. 



\begin{figure*}[t]
    \centering         
    \begin{tabular}{ccc}
    t=6 & t=12 & t=18\\
    \includegraphics[width=0.32\linewidth]{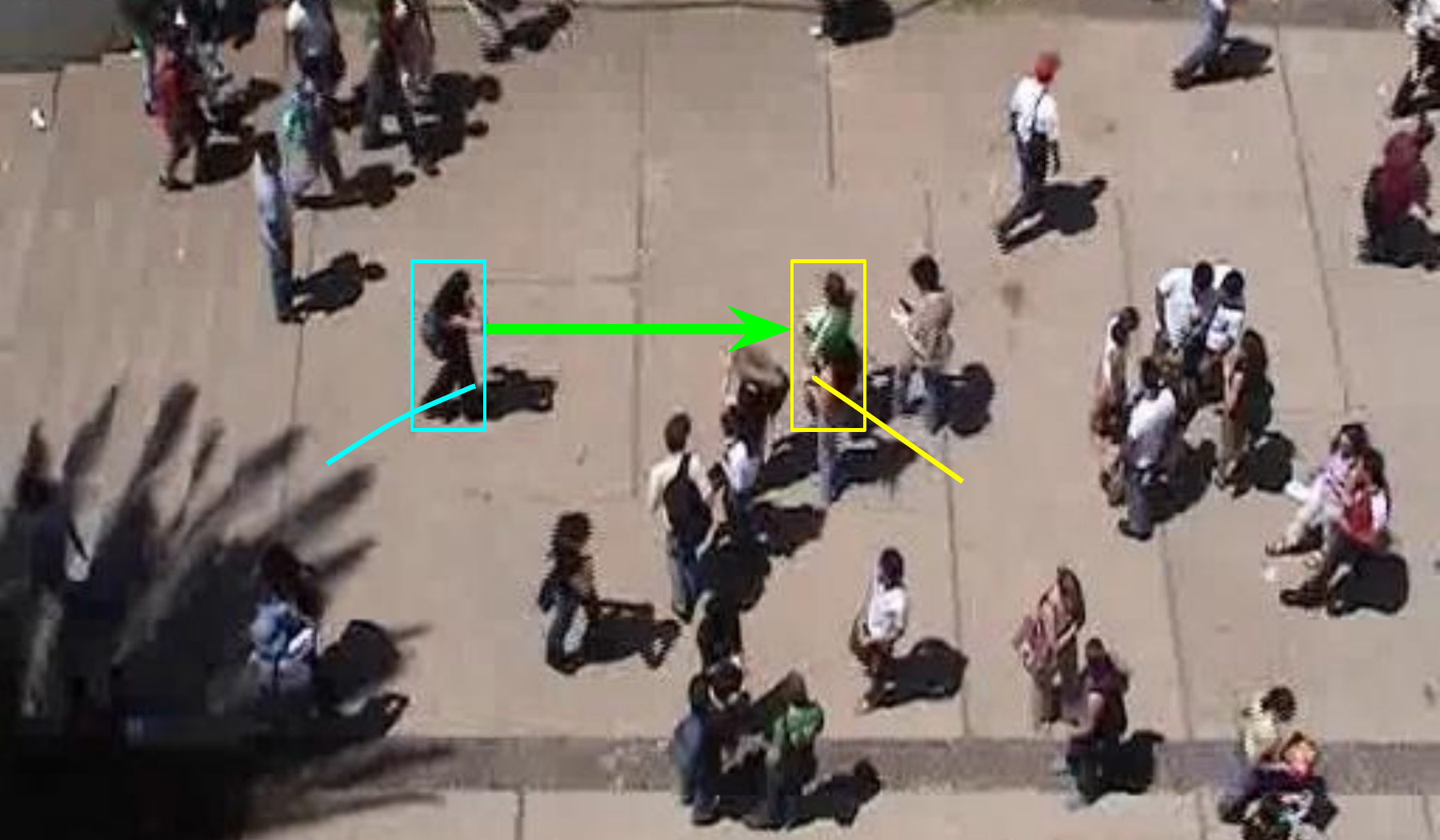}  &   
    \includegraphics[width=0.32\linewidth]{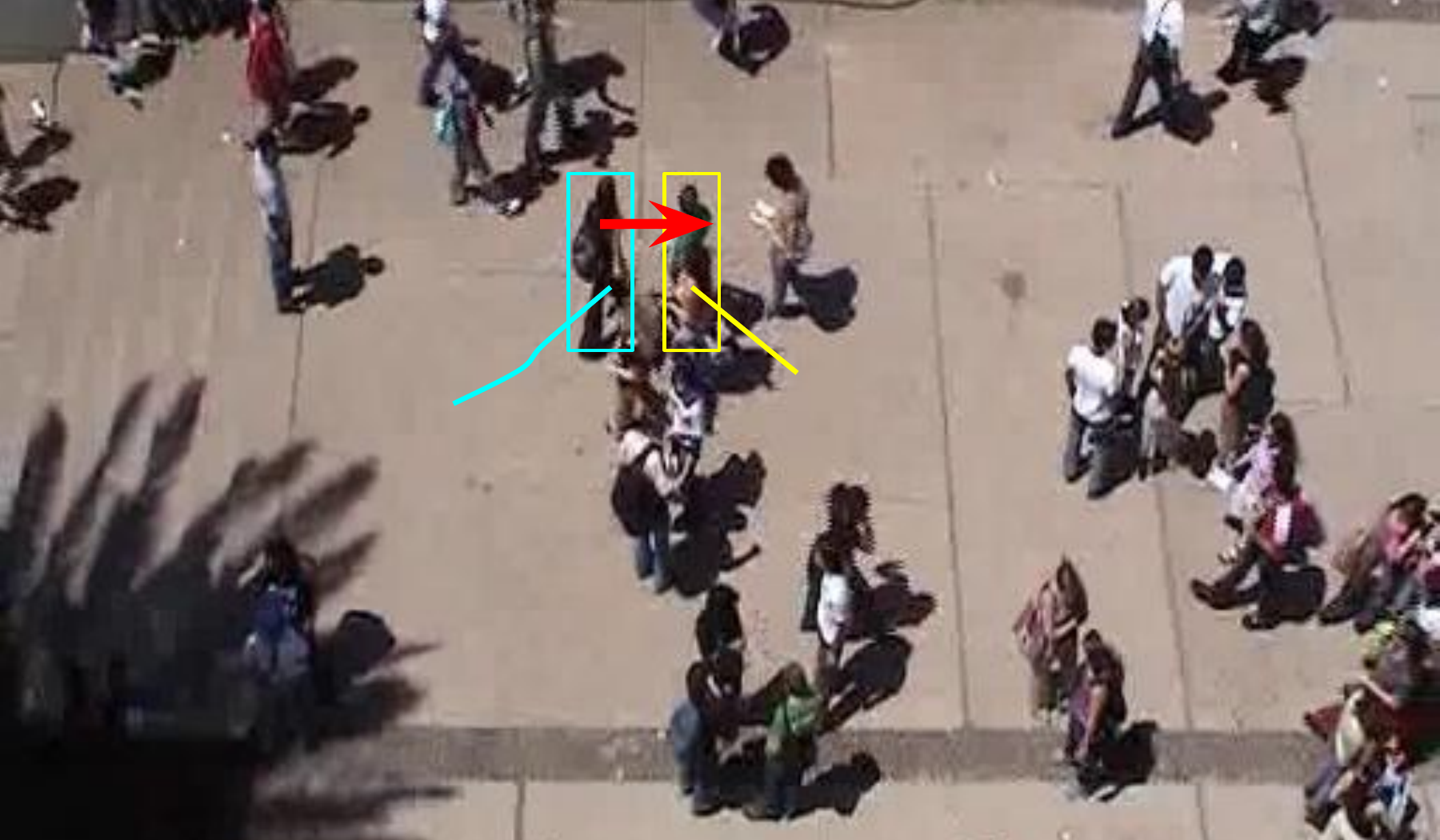} &
    \includegraphics[width=0.32\linewidth]{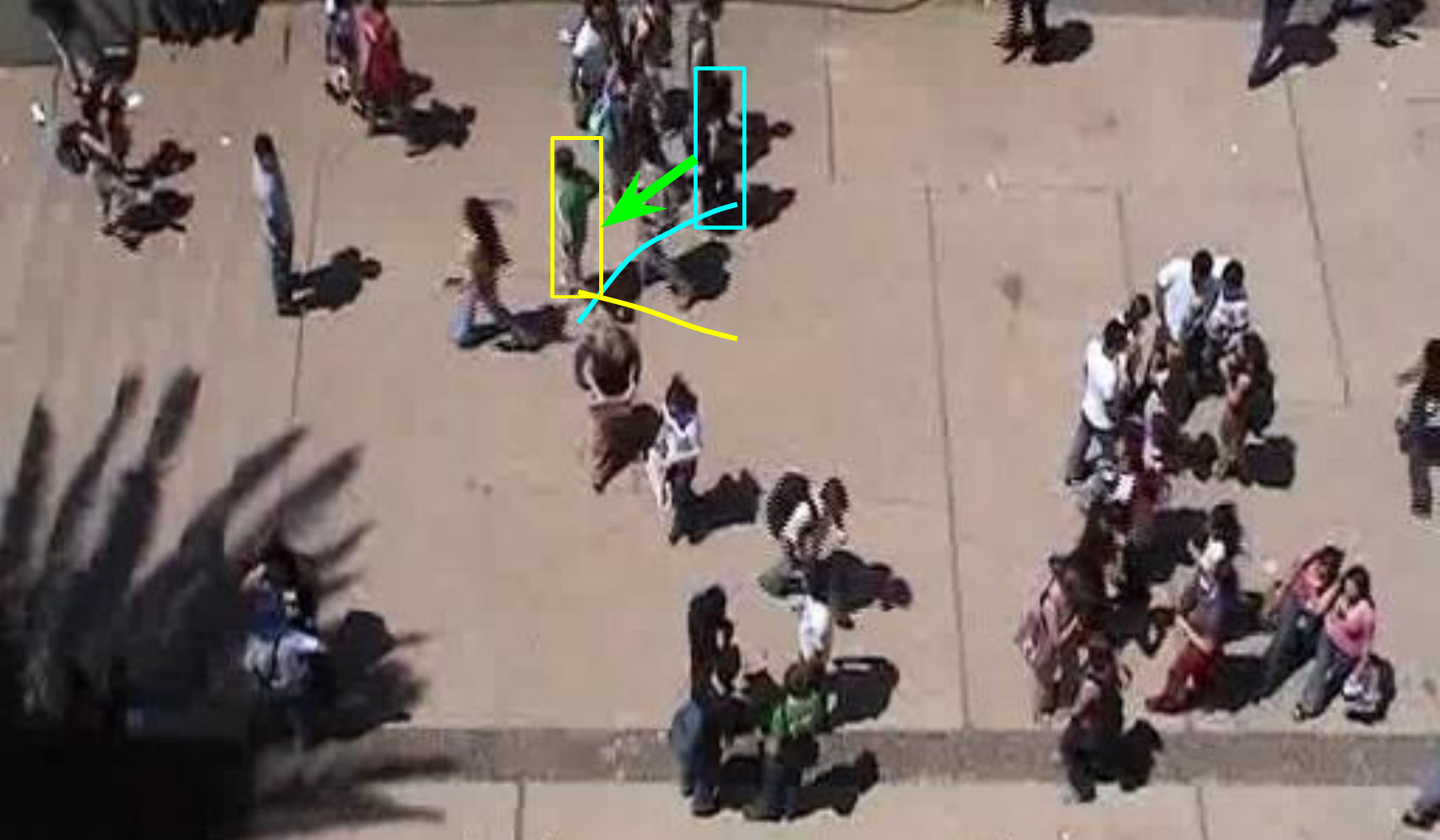} 
    \end{tabular}
    \caption{Socio-temporal dynamics: the flip of edge values from $0$ to $1$ and $1$ to $0$ indicate the occurrence of ``approaching" and ``diverging", respectively. For example, the initial edge value between two agents at $t=6$ is learned to be $0$, indicating they are uncorrelated. Afterwards, at $t=12$, when they are ``approaching" each other close enough, the edge value flips from $0$ to $1$. At time $t=18$, when they ``diverge" from each other far enough, the corresponding edge value flips from $1$ back to $0$.
    }
    \label{fig:stat_example}
\end{figure*}

\subsection{Analysis of Learned STGs}
What differentiates STGformer from other models is that past trajectories of copresent agents in the same scene are explicitly used by our model to predict any agent's future trajectory. We validate the usefulness of this information by examining its distribution in space and time, and its socio-temporal dynamics. On SDD dataset, we compute the statistics of the edge weights of each learned instance of $G^t$ and examine the relevant properties.

\bfsection{Spatial analysis:}
Fig.~\ref{fig:stat}{\red{a}} depicts the likelihood of the existence of edges of a value $1$ over space. The plot shows one agent is more likely to be correlated with her/his nearby agents; the further away two agents are located, the less likely they are to be correlated. 

\bfsection{Temporal analysis:}  
The samples are re-grouped in terms of the duration that each edge spans. The distribution presented in Fig.~\ref{fig:stat}{\red{b}} indicates two facts:(1) there are social edges that are learned to be non-trivial, indicating the socio-temporal notion pair-wise relationship between two agents can be captured through learning; (2) this notion becomes more apparent as the relationship is established in a shorter time-span. The plot shows that ``Short-term'' edges are more likely to be considered than their ``long-term'' counterparts for trajectory predictions.

\bfsection{Socio-temporal dynamics:}
We also study how the pair-wise correlations change over time and space. To this end, we define two events: agent $i$ and agent $j$ are said to be ``approaching'' each other when the corresponding distance is decreasing: $||\hat{x}^{t}_i - \hat{x}^{t}_j||_2 < ||\hat{x}^{t-1}_i - \hat{x}^{t-1}_j||_2$, or otherwise they are ``diverging'' from each other ($||\hat{x}^{t}_i - \hat{x}^{t}_j||_2 > ||\hat{x}^{t-1}_i - \hat{x}^{t-1}_j||_2$). The scenario is exemplified in Fig. \ref{fig:stat_example} 

Fig.~\ref{fig:stat}{\red{c}} shows that statistically, most edge values flip from $0$ to $1$ when two agents ``approaching'' each other, and vice-versa, most values flip from $1$ to $0$ when two agents are walking away from each other.  Moreover, the symmetry and heavy tail distribution in Fig.~\ref{fig:stat}{\red{d}} suggest that there exists a single underlying localized perimeter surrounding each agent, where when another agent enters or leaves it, the corresponding edge flips its value accordingly. 

\begin{figure*}[t]
\centering
\HS\HS\HS\HS\HS\HS\HS\HS\bf{ETH}\\
\includegraphics[width=0.19\linewidth]{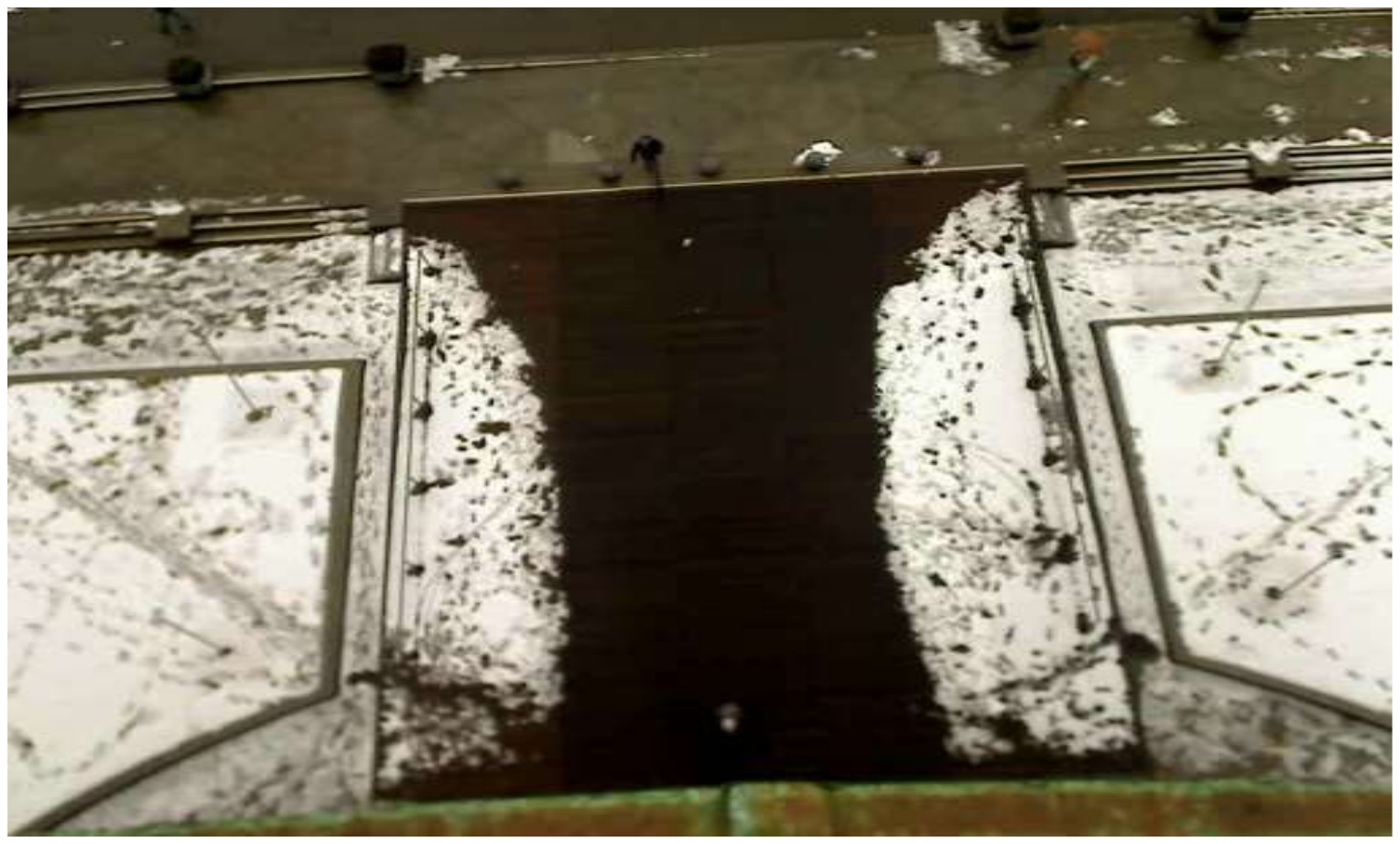}
\includegraphics[width=0.19\linewidth]{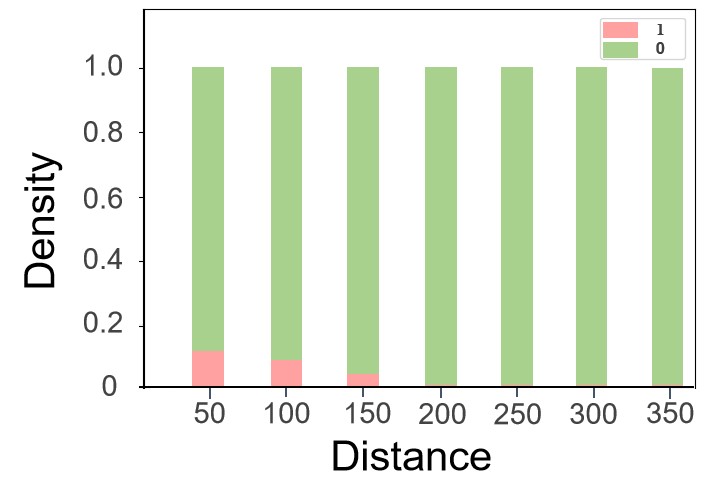}
\includegraphics[width=0.19\linewidth]{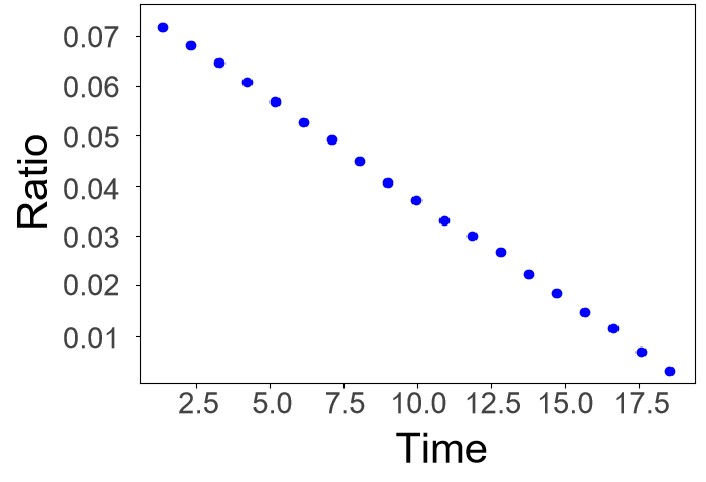} 
\includegraphics[width=0.18\linewidth]{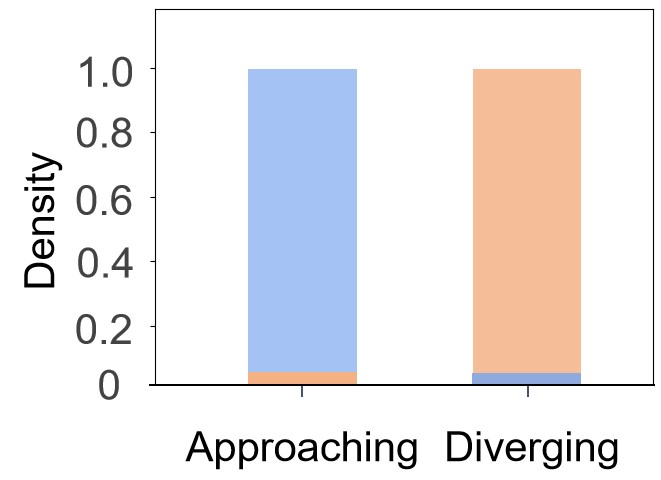}
\includegraphics[width=0.18\linewidth]{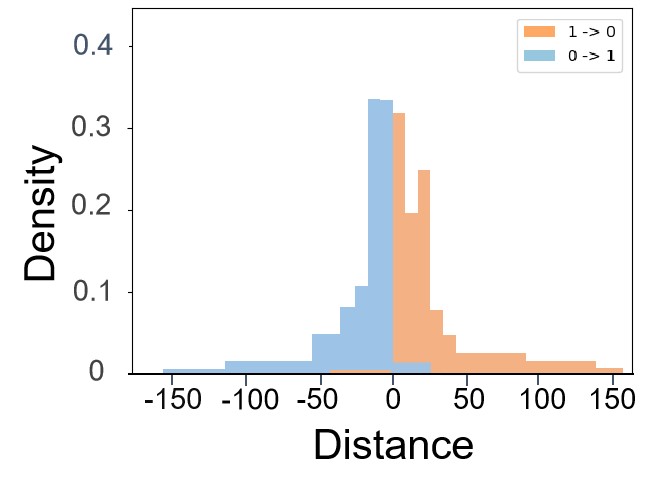}\\
\HS\HS\HS\HS\HS\HS\HS\HS\bf{Hotel}\\
\includegraphics[width=0.19\linewidth]{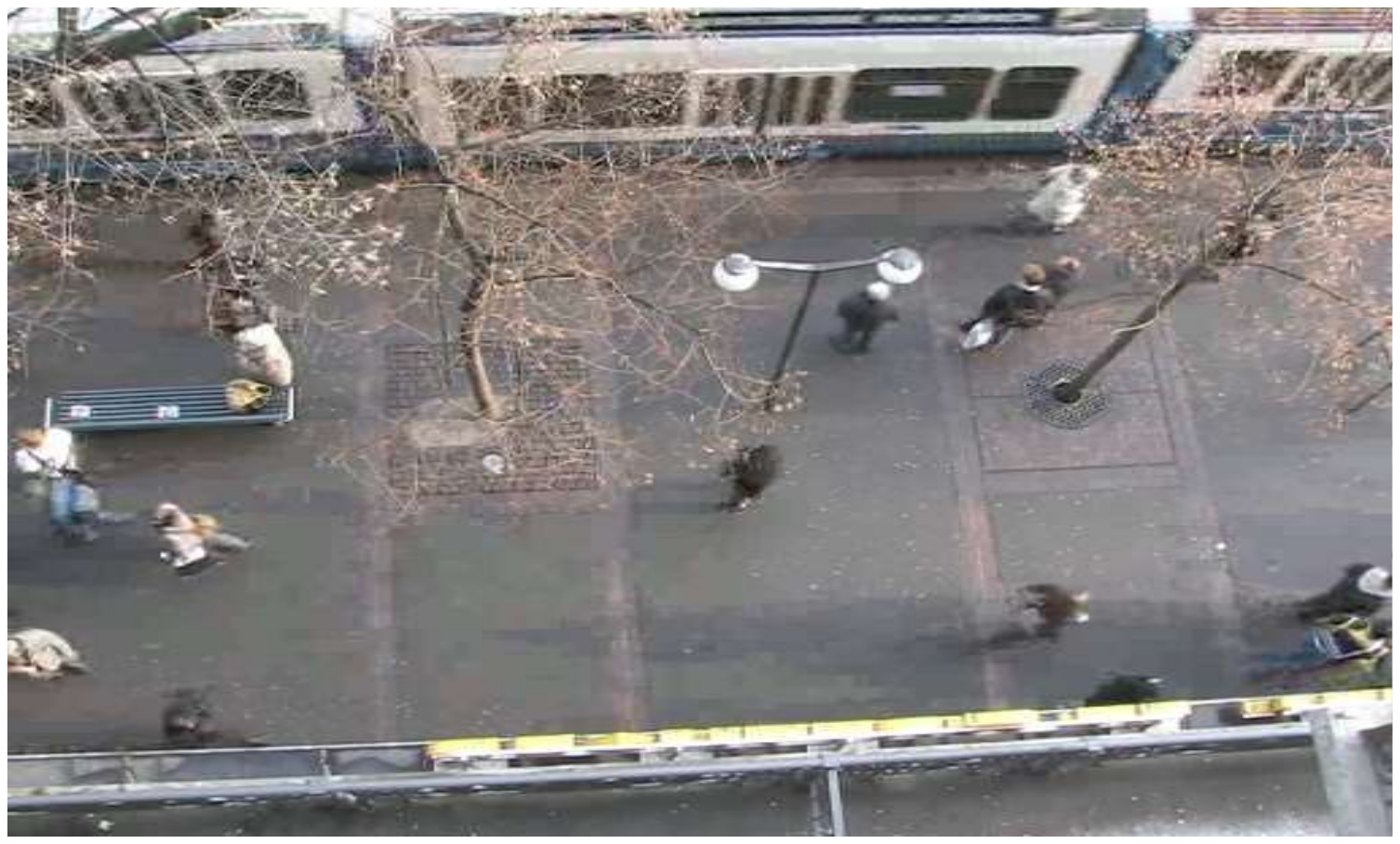}
\includegraphics[width=0.19\linewidth]{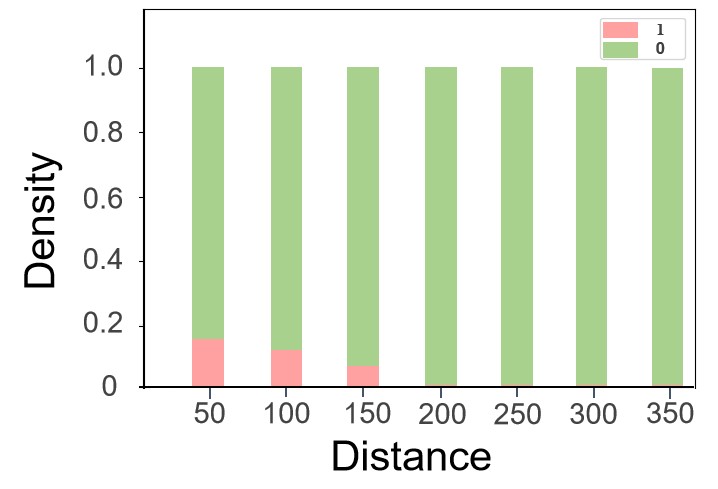}
\includegraphics[width=0.19\linewidth]{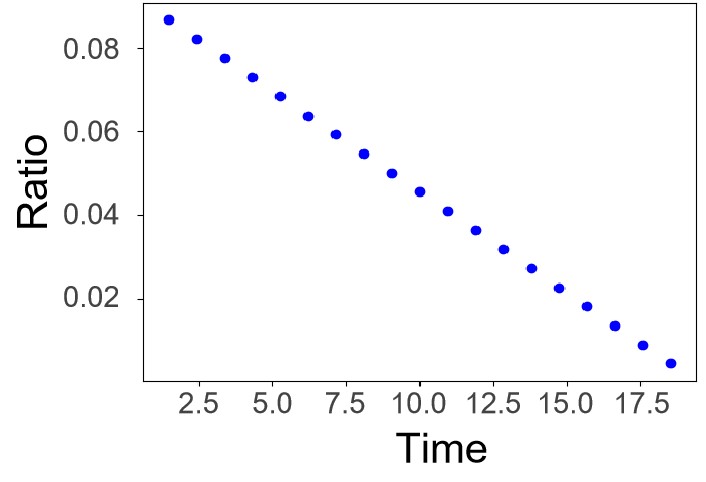} 
\includegraphics[width=0.18\linewidth]{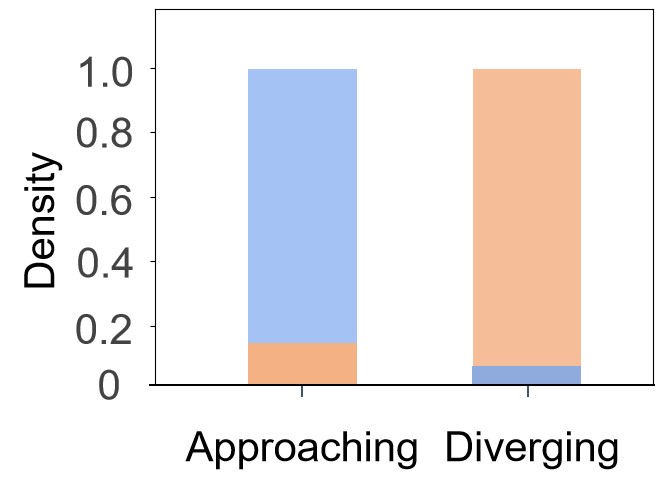}
\includegraphics[width=0.18\linewidth]{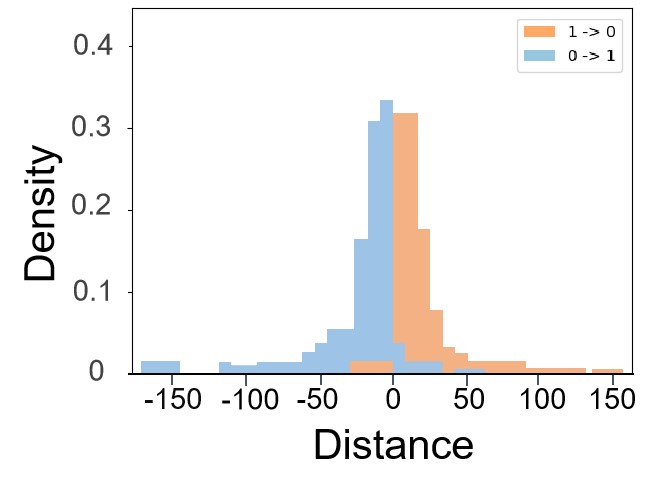}\\
\HS\HS\HS\HS\HS\HS\HS\HS\bf{Univ}\\
\includegraphics[width=0.19\linewidth]{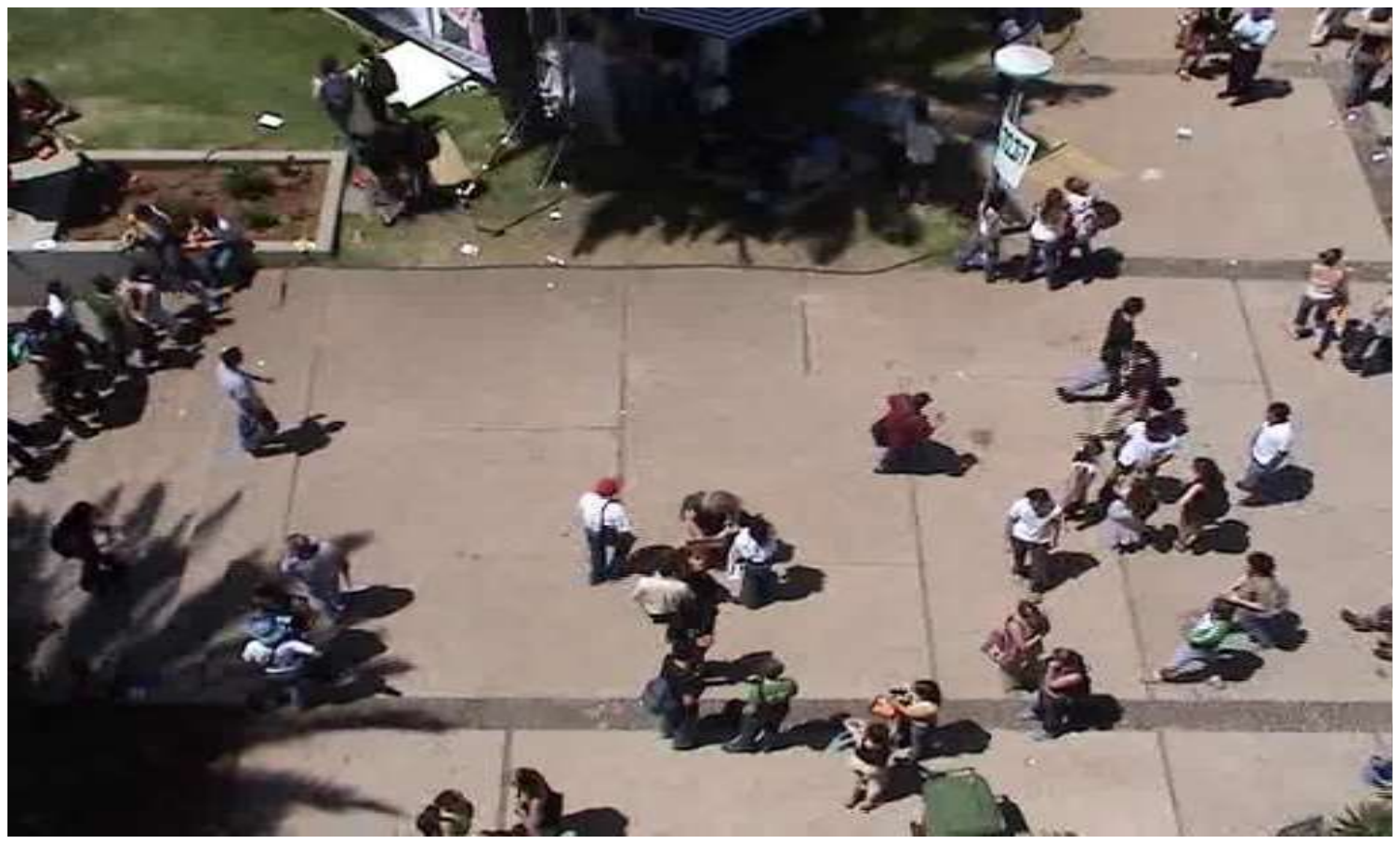}
\includegraphics[width=0.19\linewidth]{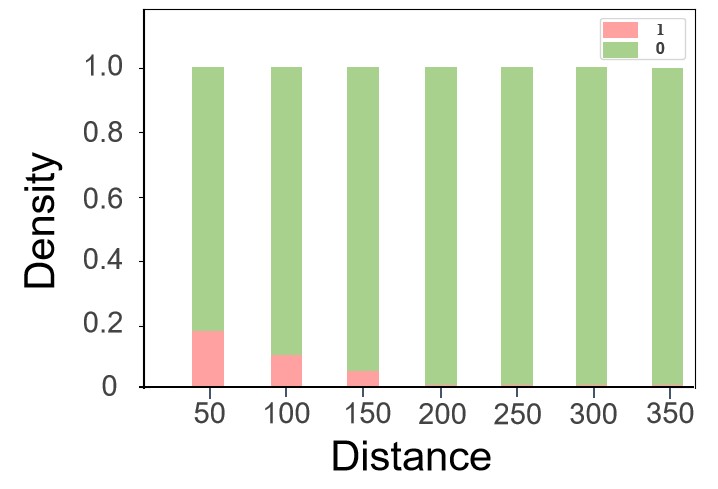}
\includegraphics[width=0.19\linewidth]{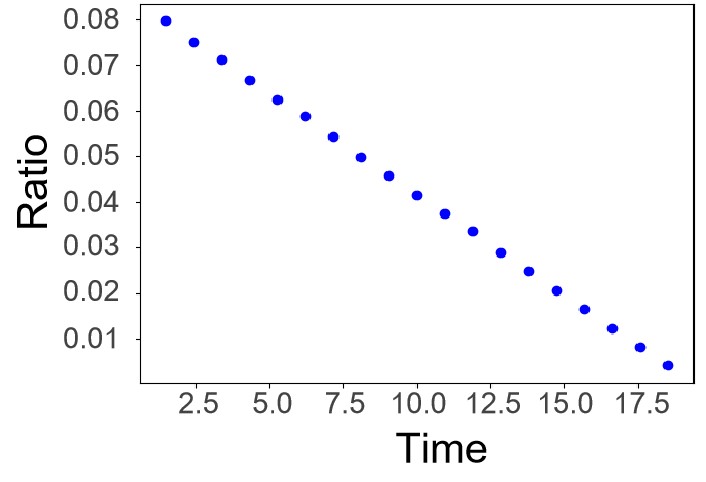} 
\includegraphics[width=0.18\linewidth]{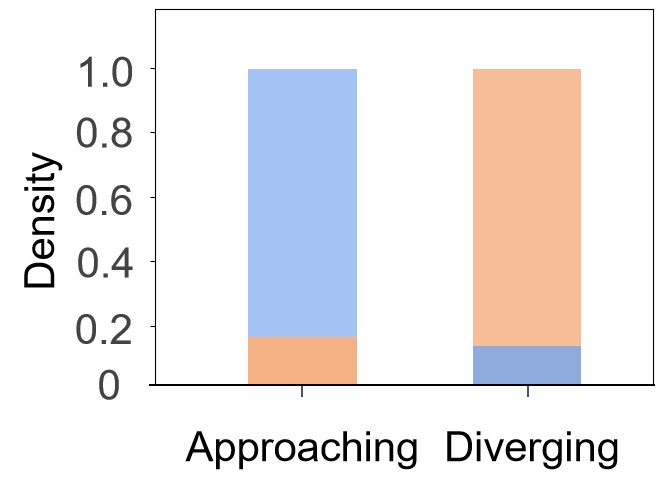}
\includegraphics[width=0.18\linewidth]{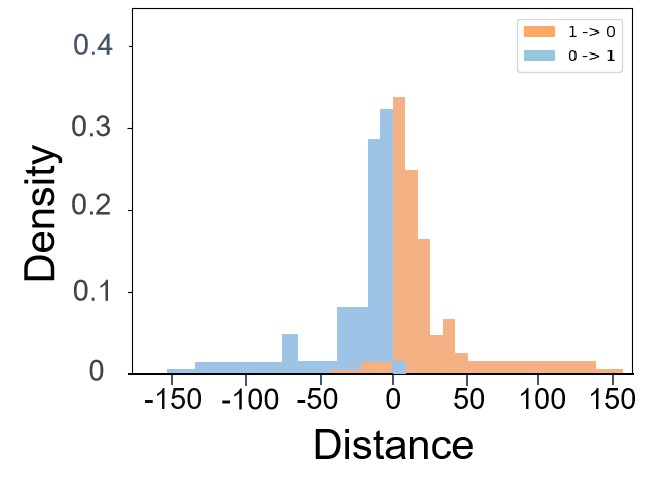}\\
\HS\HS\HS\HS\HS\HS\HS\HS\bf{Zara1}\\
\includegraphics[width=0.19\linewidth]{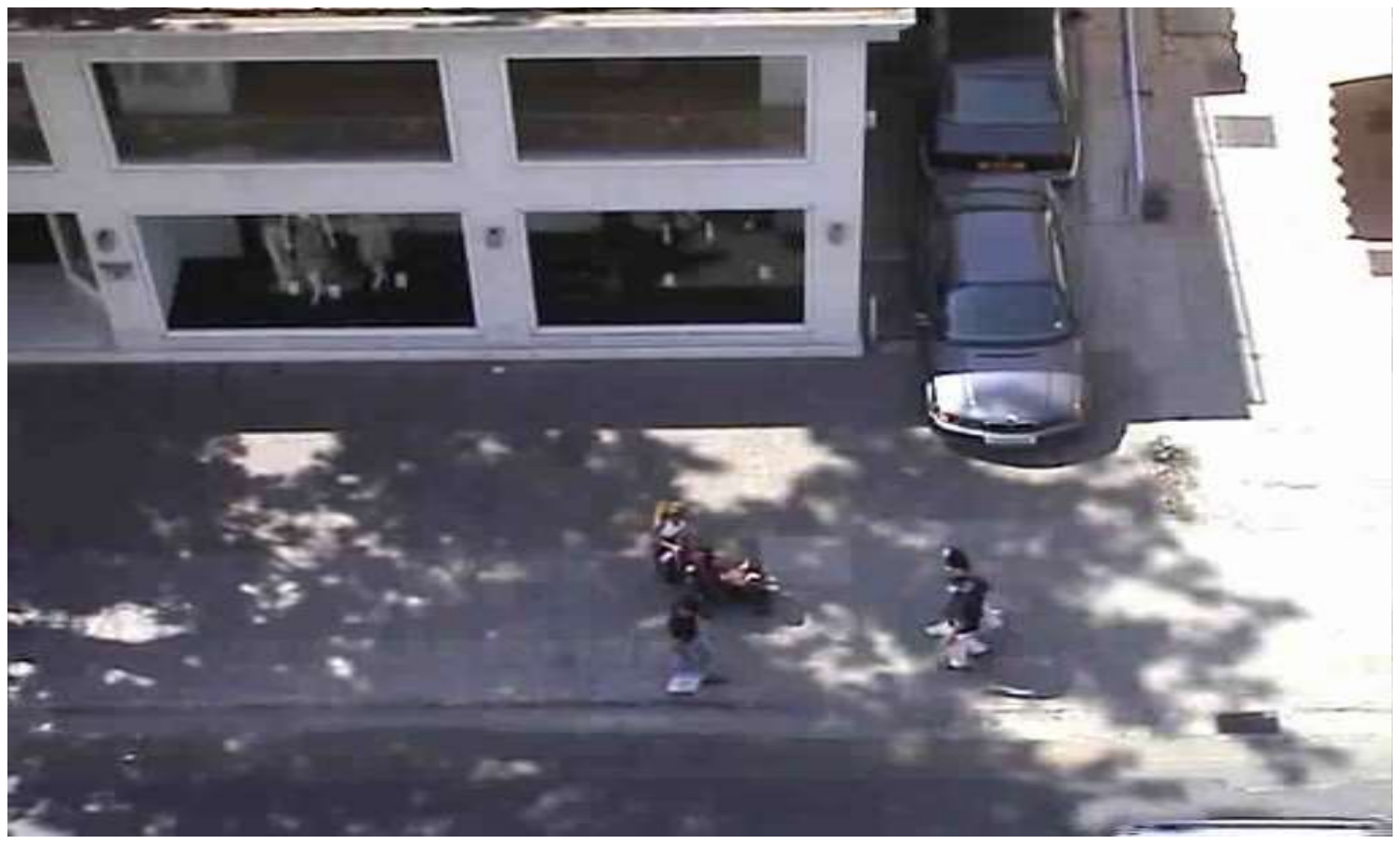}
\includegraphics[width=0.19\linewidth]{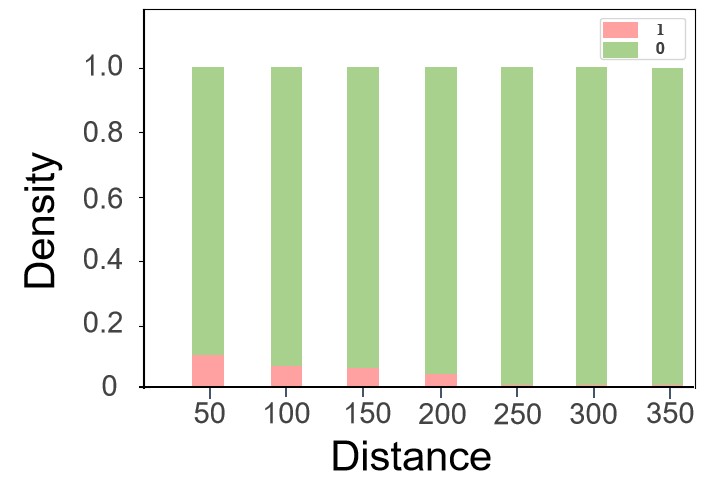}
\includegraphics[width=0.19\linewidth]{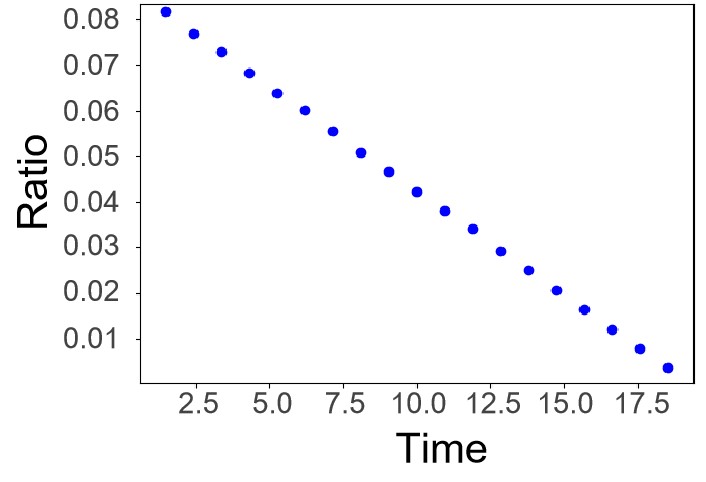} 
\includegraphics[width=0.18\linewidth]{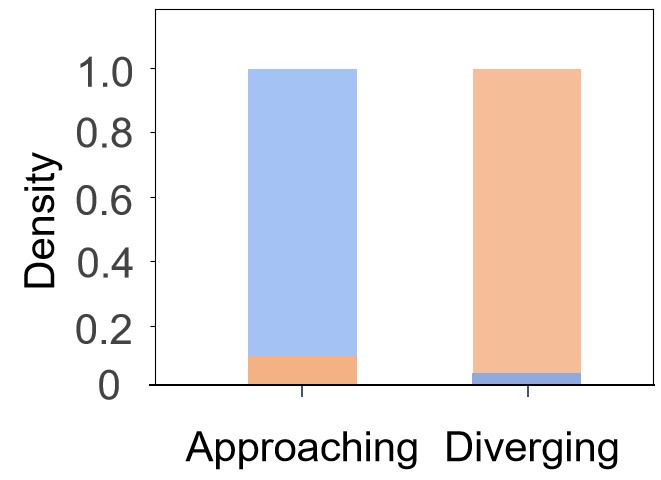}
\includegraphics[width=0.18\linewidth]{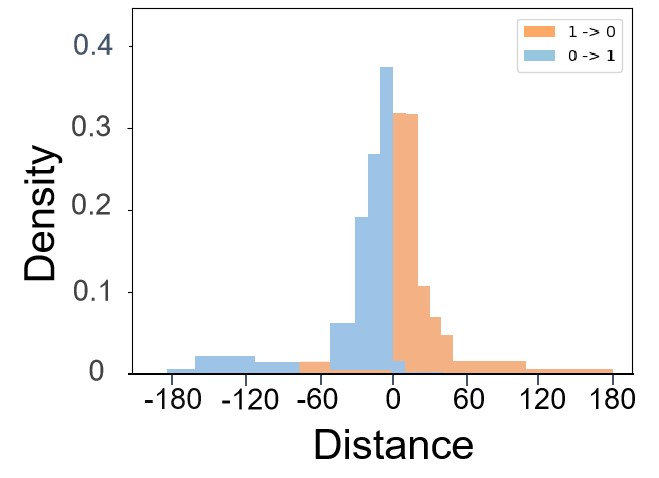}\\
\HS\HS\HS\HS\HS\HS\HS\HS\bf{Zara2}\\
\includegraphics[width=0.19\linewidth]{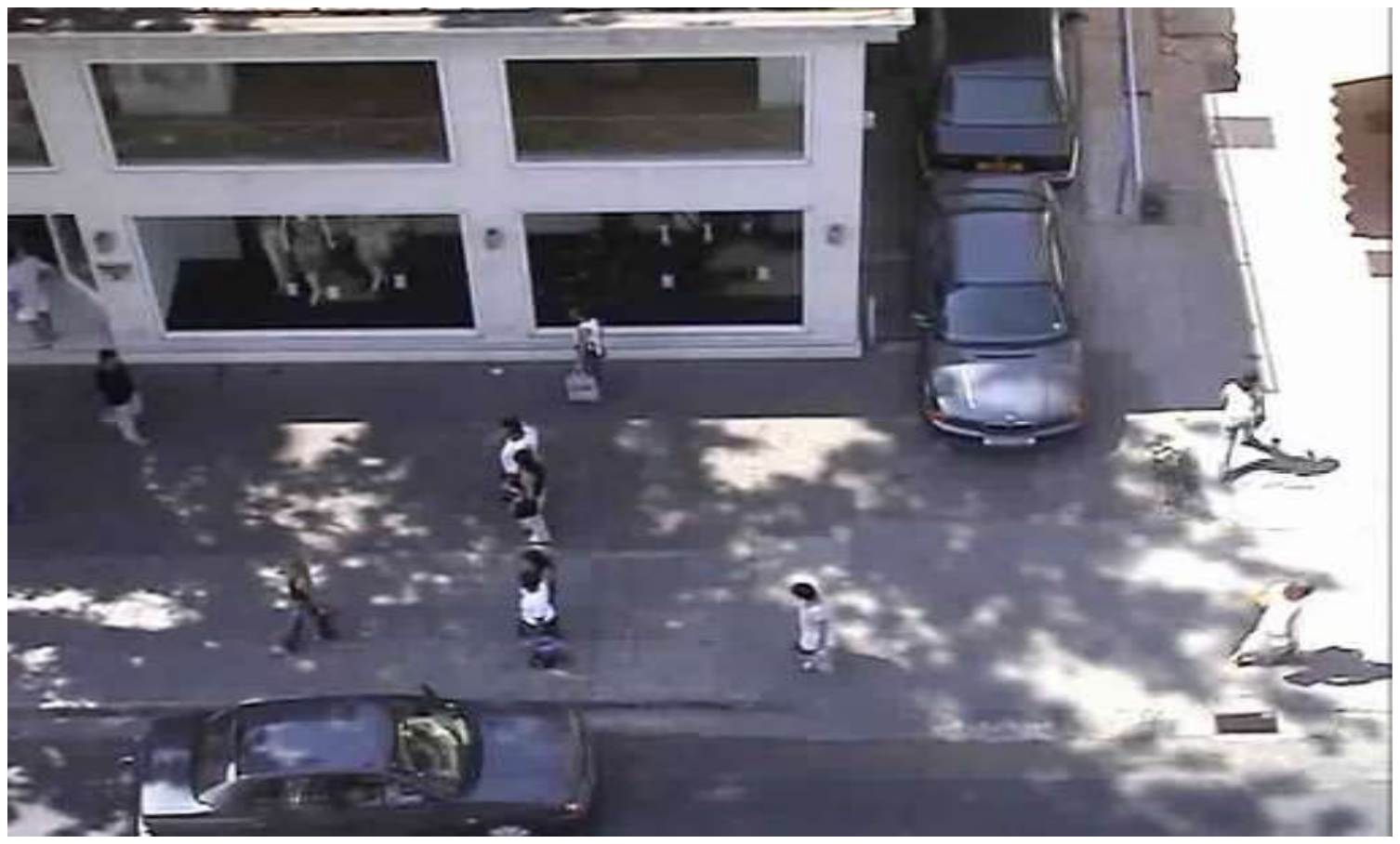}
\includegraphics[width=0.19\linewidth]{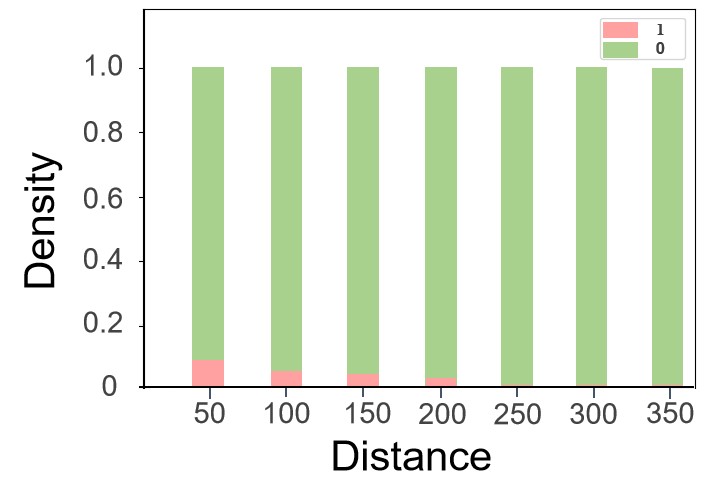}
\includegraphics[width=0.19\linewidth]{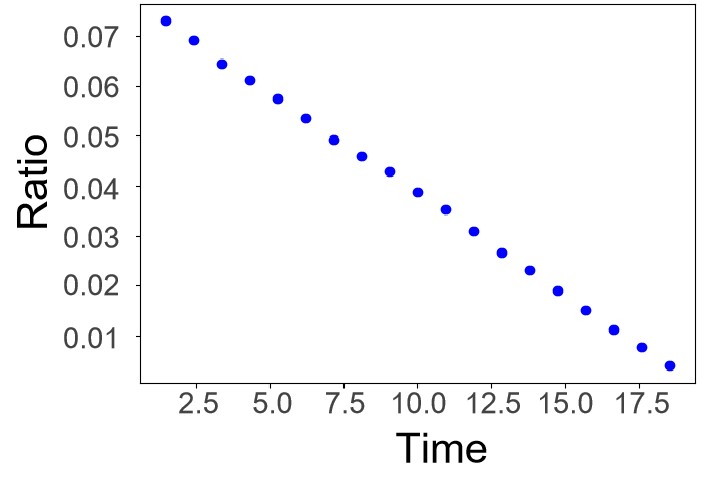} 
\includegraphics[width=0.18\linewidth]{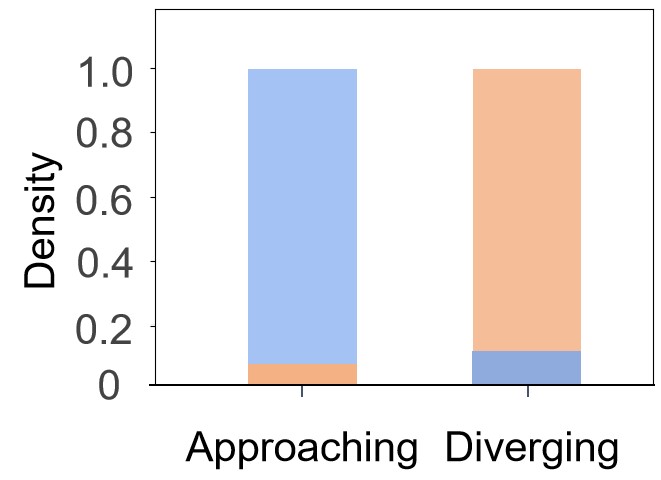}
\includegraphics[width=0.18\linewidth]{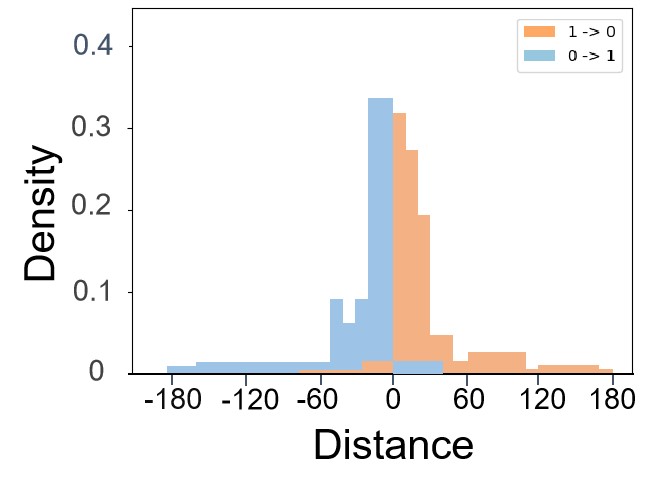}\\
\text{\HS\HS\HS\HS\HS\HS
(A) \HS\HS\HS\HS\HS\HS\HS\HS\HS\HS\HS\HS\HS\HS\HS\HS\HS\HS\HS\HS\HS\HS\HS\HS\HS
(B) \HS\HS\HS\HS\HS\HS\HS\HS\HS\HS\HS\HS\HS\HS\HS\HS\HS\HS\HS\HS\HS\HS
(C) \HS\HS\HS\HS\HS\HS\HS\HS\HS\HS\HS\HS\HS\HS\HS\HS\HS\HS\HS\HS\HS
(D)
\HS\HS\HS\HS\HS\HS\HS\HS\HS\HS\HS\HS\HS\HS\HS\HS\HS\HS\HS\HS\HS
(E)
}
\caption{
We compute the statistics of the learned edge values and study the socio-temporal notion of the pair-wise agent interactions these edges encode over the ETH/UCY dataset. (A) The different scenario settings in ETH/UCY dataset. (B) The sparse distribution of learned non-trivial STG edge values over the distance between the two agents the edges connect. Edges with a learned value of $1$ mostly link two agents no more than $ 200$ pixels, meaning the behaviors of most agents outside this range are deemed less essential for trajectory predictions. (C) The distribution of edges with a value of $1$ over time. ``Short-term'' edges are more likely to be considered than their ``long-term'' counterparts for trajectory predictions. (D) The flip of edge from $0$ to $1$ is highly correlated with the event where two agents  ``approaching" each other, and vice-versa. The event that two agents ``diverging" from each other is captured by the edge value flipping from $1$ to $0$. (E) Symmetry and spatial localization suggest that most events occur when one agent enters or leaves another agent around a $60$-pixel-perimeter.}
\label{fig:addtional}
\end{figure*}

\bfsection{Analysis of Learned STGs on ETH/UCY datasets}
We also conduct the same analysis of the learned STGs on each scene of the ETH/UCY datasets (Fig.~\ref{fig:addtional}\red{A}). The learned STGs on the ETH/UCY datasets have similar characteristics to that on the SDD dataset. Specifically, Fig.~\ref{fig:addtional}\red{B} 
reveals the likelihood of the existence of edges of a value 1 over space.
The distribution presented in Fig.~\ref{fig:addtional}\red{C} indicates two facts:(1) the socio-temporal notion pair-wise relationship between two agents can be captured through learning STGs; (2) “Short-term” edges are more likely to be considered than their “long-term” counterparts for trajectory predictions.
Fig.~\ref{fig:addtional}\red{D} illustrates that statistically, most edge values flip from 0 to 1 when two agents “approaching” each other, and vice-versa, most values flip from 1 to 0 when two agents are walking away from each other.  
The symmetry and heavy tail distribution in Fig.~\ref{fig:addtional}\red{E} suggest that there exists a single underlying localized perimeter surrounding each agent, where when another agent enters or leaves it, the corresponding edge flips its value accordingly.
 
\subsection{Ablation Study}
We conducted the following ablation studies to examine the effectiveness of the model design.

\begin{table*}[t]
\begin{center} 
\caption{Ablation studies on the ETH/UCY and SDD datasets. We report the ADE / FDE scores over 20 predictions.} 
\vspace{-0.2cm}
\begin{tabular}{ll|ccccc|c|c}
\hline
&  &
{ETH}  & {Hotel} &  { Univ.}&{ Zara1}&{ Zara2}&{ AVG} &{ SDD} \\ \hline
w/o social
& &  0.79/1.12  & 0.61/0.74 & 0.53/0.85 &  0.69/0.92 & 0.62/1.07 & 0.65/0.94 & 14.71/21.07
\\
w/o LP
& &  0.62/1.07  & 0.49/0.55 & 0.45/0.69 &  0.42/0.61 & 0.47/0.66 & 0.49/0.72 & 12.93/17.42
\\
Stationary-$G^t$
& & 0.55/0.84  & 0.44/0.49 & 0.44/0.65 &  0.38/0.60 & 0.34/0.52 & 0.43/0.62 & 11.40/17.15\\
Short-term-$\mathcal{G}^t$
& & 0.45/0.69 & 0.31/0.42 & 0.35/0.63 & 0.30/0.52 & 0.31/0.38 & 0.34/0.45 & 9.04/14.29\\
w/o $G^t$
& & 0.44/0.65  & 0.26/0.38 & 0.31/0.59 & 0.23/0.47 & 0.22/0.35 & 0.29/0.49 & 8.83/14.61\\
$\zeta=0$
& & 0.32/0.60  & 0.24/0.33 & 0.25/0.51 & 0.21/0.38 & 0.18/0.27 & 0.24/0.42 & 8.28/12.90\\
\hline
\textbf{\model~(Ours)}
& & \bf{0.27}/0.56  & \bf0.11/0.17 & \bf{0.22/0.45} & \bf{0.16}/0.31  & \bf{0.14/0.24}  & \bf{0.18}/0.35  & \bf7.35/{11.39} \\
\hline
\end{tabular}
\label{tab:ablation}
\end{center}
\vspace{-0.4cm}
\end{table*}

\begin{table*}[t]
\begin{center} 
\caption{Ablation studies on the architectures of the ADE / FDE scores over 20 predictions on the ETH and UCY datasets. The last row is our STGFormer results.} 
\vspace{-0.2cm}
\begin{tabular}{cc|ccccc|c|c}
\hline
$G^t$ & $\itx^t$ &
{ETH}  & { Hotel} &  { Univ.}&{ Zara1}&{ Zara2}&{ AVG} &{ SDD}\\ \hline
LSTM
& GAT & 0.55/0.76  & 0.49/0.64 & 0.50/0.65 &  0.38/0.54 & 0.42/0.57 & 0.47/0.63 & 12.71/19.84\\
LSTM
& Transformer & 0.40/0.62  & 0.31/0.38 & 0.36/0.54 &  0.32/0.50 & 0.30/0.42 & 0.34/0.45 & 9.55/14.10\\
 Transformer
& GAT & 0.31/0.61  & 0.22/0.25 & 0.28/0.48 & 0.19/0.41 & 0.20/0.36 & 0.24/0.41 & 8.12/13.27\\
\hline
\bf{Transformer}
& \bf{Transformer}
& \bf{0.27}/0.56  & \bf0.11/0.17 & \bf{0.22/0.45} & \bf{0.16}/0.31  & \bf{0.14/0.24}  & \bf{0.18}/0.35  & \bf7.35/{11.39}  \\
\hline
\end{tabular}
\label{tab:architecture}
\end{center}
\vspace{-0.6cm}
\end{table*}

\bfsection{Learning socio-temporal correlations over time: w/o $G^t$} To highlight the advantages of learning the socio-temporal dependencies, we compared two models, one with $G^t$ and one without it. A design with a predefined w/o $G^t$ baseline is similar to a transformer decoder \cite{gpt2}. Again, a significant contribution of our work is to allow this entity to be learnable. We conduct ablation studies to show the benefit of learning data-driven STG over fixed ones for trajectory predictions (Tab.~\ref{tab:ablation}). This suggests that understanding socio-temporal correlations enables STGformer to generalize over various datasets over a more stable distribution.

{\bfsection{Long-term v.s. short-term correlations: Short-term-{$\mathcal{G}^t$}}} To verify that the information carried by non-trivial edges facilitates trajectory prediction, we also intentionally mask off this information in our ablation experiment. The result is reported in Tab.~\ref{tab:ablation}, annotated as ``w/o social''. It can be easily seen that if relevant information learned by our model is intentionally ignored, the performance of our model drops, which means that the information uniquely learned by our model can be used to improve prediction accuracy.

Moreover, we divide all edges of $\mathcal{G}^t$ by the time duration they span into ``short-term'' and ``long-term'', respectively. Tab. \ref{tab:ablation} reports the results if the samples only from the same group are used for prediction. The results show that the deeper we look back into history to consider potential correlations between past observations with present ones, the better prediction we make, and reasonably our performance gain becomes marginal as excessively old observations tend to carry obsolete information.  

Specifically, we adjust the duration of the ``look-back'' window of $\mathcal{G}^t$ from $\mathbb{R}^{\textit{n}\times\textit{n}t}$ to $\mathbb{R}^{\textit{n}\times\textit{n}}$.
Such a design suggests connecting an individual with his/her cause persons at only 1 step before, respectively. Our approach drastically advances short-tem $\mathcal{G}^t$ by improving from $9.04/14.29$ to $7.35,11.39$ on SDD, and from $0.34/0.45$ to $0.18,0.35$ on ETH/UCY. The results suggest that the long-term correlations provide valuable information for future predictions. 
    
\bfsection{Time varying property: Stationary-$G^t$}
The performance of STGFormer is considerably better than its counterpart design with a learned-once-applied-to-all stationary-$G^t$ baseline, which assumes that the system is time-invariant. Under the time-invariant assumption, $G^t$ at the first step is learned and applied to analyze all subsequent observations. The results suggest that the system governing human trajectories is time-varying.


\bfsection{Generating $G^t$: w/o LP (learned prior)} 
We compare models with and without LP baseline. More specifically, the model without LP drops $\itp_\Psi(G^{t}|G^{0:t-1})$ by setting $G^t \sim \mathcal{N}(0,\textit{I})$. 
Tab.~\ref{tab:ablation} illustrates the advantage of \modelo over the model without LP, indicating the necessity of capturing the underlying temporal correlation of $G^t$ over $t$ upon learning from $p_{\Psi}(G^{t}|G^{0:t-1})$.

\bfsection{STG sparsity term: $\zeta=0$}
The sparsity term means $\zeta||\mathcal{G}^{0:T}||_0$. We favor a succinct socio-temporal interpretation for our observation. Setting $\zeta=0$ allows us to minimize the weight of edges encoding unnecessary correlations. We conducted an ablation study to verify this point that trains the model without $\zeta||\mathcal{G}^{0:T}||_0$. Tab.~\ref{tab:ablation} shows that with the sparsity regularizer, the learner can generate better/more succinct interpretations, which lead to more accurate predictions by our design.

\bfsection{Transformer Architecture} 
In order to further verify the effectiveness of our architecture choices, we tested models with a different backbone, choosing LSTM \cite{bptt_2016} and GAT \cite{gat_iclr18} as alternatives in contrast to the transformer. 
To ensure a fair comparison, we used a single LSTM layer with a hidden size of 256 to learn the prior $p_{\Psi}(G^{t}|G^{0:t-1})$ and posterior $q_{\Theta}(G^{t}|G^{0:t-1},\bfx^{0:t})$. The trajectory model, implemented with a Two-stacked GAT, is set to a size of 512. Both GAT layers use eight attention heads. 


\section{Conclusion}\label{sec:clu}
In this paper, we introduce and test a \model~for learning to better foresee human trajectories. 
The \model~ models the joint distributions that are formulated in ~Eqn.~\ref{eq:overall}. This formulation allows us to grasp the socio-temporal graph structures underlying representations of human trajectories. Experiment results demonstrate that our model delivers better performance in the task of path forecasting compared with other state-of-the-art trajectory-based approaches.
We believe that our \model~can paves the way for future improvements. One promising direction is to extend our framework to include context information, such as images, and trajectories, to discover the correlations between human agents and the environment in terms of socio-temporal interactions.



\bibliographystyle{IEEEtran}
\bibliography{egbib}

\newpage

 




\vfill

\end{document}